\documentclass{article}

\usepackage{PRIMEarxiv}

\usepackage[utf8]{inputenc} 
\usepackage[T1]{fontenc}    
\usepackage{hyperref}       
\usepackage{url}            
\usepackage{booktabs}       
\usepackage{amsfonts}       
\usepackage{nicefrac}       
\usepackage{microtype}      
\usepackage{lipsum}
\usepackage{fancyhdr}       
\usepackage{graphicx}       
\usepackage{amsmath}
\usepackage{xcolor}
\usepackage{url}
\usepackage{multirow}
\usepackage{booktabs}
\usepackage{caption}
\usepackage{subcaption}
\usepackage{adjustbox}
\usepackage{hyperref}
\usepackage{algorithm}
\usepackage{algorithmic}
\usepackage{longtable}
\usepackage{pdflscape}   
\usepackage{booktabs}   
\usepackage{multirow}
\usepackage{subcaption}
\usepackage[numbers]{natbib}
\newcommand{\ignore}[1]{}

\title{On the Use of Bi-Objective Evolutionary Algorithms for the Stochastic Multiple Knapsack Problem under Dynamic Constraints
}

\author{\href{https://orcid.org/0000-0002-8043-994X}{\includegraphics[scale=0.06]{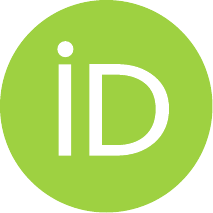}\hspace{1mm}Ishara Hewa Pathiranage} \\
	Machine Learning and Optimisation, \\School of Computer Science and \\Information Technology,\\
  Adelaide University,\\ Adelaide, Australia 
	\And
	\href{https://orcid.org/0000-0002-0036-4782}{\includegraphics[scale=0.06]{orcid.pdf}\hspace{1mm}Aneta Neumann} \\
	Machine Learning and Optimisation, \\
    School of Computer Science and\\ Information Technology,\\
  Adelaide University,\\ 
  Adelaide, Australia \\
}

\author{
\href{https://orcid.org/0000-0002-8043-994X}{\includegraphics[scale=0.06]{orcid.pdf}}
\hspace{1mm}Ishara Hewa Pathiranage \\
Machine Learning and Optimisation, \\School of Computer Science and \\Information Technology,\\
Adelaide University,\\ Adelaide, Australia 
\And
\href{https://orcid.org/0000-0002-0036-4782}{\includegraphics[scale=0.06]{orcid.pdf}}
\hspace{1mm}Aneta Neumann \\
Machine Learning and Optimisation, \\
School of Computer Science and\\ Information Technology,\\
Adelaide University,\\ 
Adelaide, Australia \\
}

\begin{document}
\maketitle

\begin{abstract}
The multiple knapsack problem~(MKP) generalizes the classical knapsack problem by assigning items to multiple knapsacks subject to capacity constraints. It is used to model many real-world resource allocation and scheduling problems. In practice, these optimization problems often involve stochastic and dynamic components. Evolutionary algorithms provide a flexible framework for addressing such problems under uncertainty and dynamic changes. In this paper, we investigate a stochastic and dynamic variant of MKP with chance constraints, where the item weights are modeled as independent normally distributed random variables and knapsack capacities change during the optimization process. We formulate the problem as a bi-objective optimization formulation that balances profit maximization and probabilistic capacity satisfaction at a given confidence level. We conduct an empirical comparison of four widely used multi-objective evolutionary algorithms~(MOEAs), representing both decomposition-
and dominance-based search paradigms. 
The algorithms are evaluated under varying uncertainty levels, confidence thresholds, and dynamic change settings. The results provide comparative insights into the behavior of decomposition-based and dominance-based MOEAs for stochastic MKP under dynamic constraints.
\end{abstract}



\maketitle

\section{Introduction}
Many real-world optimization problems involve stochastic components due to the variations in
resources, demand, or operating conditions~\citep{10.1145/3638529.3654066}.
For example, renewable energy availability is inherently stochastic and dependent on weather conditions,
requiring power systems to balance supply and demand under uncertainty~\cite{en14248240}.
To account for such uncertainties, chance constraints~\citep{Charnes} are
integrated into optimization problems to ensure that stochastic constraints
are satisfied with high probability, usually denoted by $\alpha$. Chance-constrained optimization has been studied in various domains such as mining~\cite{doi:10.1080/25726668.2021.1916170,10.1145/3449639.3459382,10254112}, and logistics~\cite{10.5555/2888116.2888219,
FARINA201653}, where it is beneficial
to ensure that a performance measure remains above a specified threshold with
high probability rather than optimizing only its expected value.

In addition to uncertainty, many optimization problems are dynamic, where
objectives or constraints change over time~\cite{10.1145/3524495}. For instance, in mining truck
allocation, vehicle capacities may vary due to the operational and environmental
factors. Such dynamic changes increase problem complexity and motivate the use
of algorithms that can easily adapt to such conditions. Without adaptation,
solutions obtained for earlier environments may become suboptimal or infeasible
in practice~\citep{10.1145/3638529.3654067}.

The multiple knapsack problem (MKP) generalizes the
classical knapsack problem by assigning items to multiple knapsacks under
capacity constraints to maximize total profit. MKP has been widely used as a
benchmark for studying resource allocation problems, including smart home
energy management~\citep{Shewale2024}, load balancing in distributed
systems~\citep{Li2015}, and resource planning in mining
operations~\citep{SOUZA20101041,mine}. 
Comprehensive overview on MKP and its variants can be found in
\citep{CACCHIANI2022105693}. Compared to the single knapsack problem, MKP is
more challenging due to the multiple interacting capacity and reserve
constraints~\citep{MANCINI2021987}. The problem becomes significantly more
complex when the item weights are stochastic and the knapsack capacity bounds change
over time.

Evolutionary algorithms~(EAs) are bio-inspired randomized optimization techniques that provide a flexible framework for exploring large and complex
search spaces. They have been widely
used to solve various combinatorial optimization
problems~\cite{coello2007evolutionary,DBLP:conf/gecco/NeumannGYSCG023}. Moreover, they have shown effectiveness in solving dynamic and chance constrained variants~\cite{RAKSHIT201718, Neumann2020, Roostapour2020, DBLP:conf/ecai/AssimiHXN020,NGUYEN20121, myburgh2010evolutionary}. When dynamic and chance-constrained optimization problems are solved with EAs, multi-objective
evolutionary algorithms~(MOEAs) demonstrate better results than the single-objective approach~\cite{ijcai2022p665, 10.1145/3377930.3390162}. MOEAs allow to model chance constraints as additional objectives and it is beneficial to optimize different stochastic components~(e.g. mean and variance) simultaneously
\citep{ijcai2022p665,DBLP:conf/gecco/0001W23,10.1162/evco_a_00360}. Recent works have investigated the use of MOEAs for chance-constrained and dynamic variants of
several combinatorial optimization problems, including knapsack~\cite{DBLP:conf/ecai/AssimiHXN020,10.1145/3377930.3390162,10.1145/3638529.3654066,10.1145/3638529.3654067,10.1145/3638529.3654081}, submodular~\cite{10.1162/evco_a_00360}, and dominating set
problems~\citep{DBLP:conf/gecco/0001W23}. 
In this paper, we investigate a bi-objective formulation of the chance-constrained
multiple knapsack problem with normally distributed item weights and dynamic
capacity bounds. We empirically evaluate the behavior of dominance-based and
decomposition-based MOEAs under several stochastic and dynamic settings.

\subsection{Related Work}

The multiple knapsack problem is a well-known NP-hard combinatorial
optimization problem that has been extensively studied due to its relevance in many application domains.
To solve this, a wide range of exact solution methods have been proposed, including
multi-criteria dynamic programming~\citep{SITARZ2014598}, valid inequality
generation~\citep{Hickman}, pseudo-polynomial arc-flow formulations combined
with decomposition techniques~\citep{DELLAMICO2019886}, and improved upper-bound
formulations for large-scale instances~\citep{DETTI2021105210}.
While these approaches are effective for moderate problem sizes, their
scalability is often limited for large-scale MKP instances.

To overcome these limitations, numerous heuristic and meta-heuristic methods
have been developed, including greedy and surrogate relaxation
methods~\citep{laalaoui2016binary}, approximation heuristics~\citep{Chekuri2000213},
dynamic programming-based heuristics~\citep{lalami2012procedure}, and
population-based meta-heuristics such as genetic
algorithms~\citep{10.5555/1689599.1689921}, particle swarm
optimization~\citep{DBLP:conf/icnc/RenW09}, discrete evolutionary
algorithms~\citep{WANG2024121170}, and deep reinforcement learning
approaches~\citep{sur2022deep}.

Compared to the extensive literature on classical MKP, stochastic and dynamic
variants have received considerably less attention.
Early work examined solution representations for dynamic MKP~\citep{10.1007/11732242_74},
followed by genetic algorithm variants incorporating random immigrants and
memory-based mechanisms~\citep{unal2013genetic}.
More recently, chance-constrained formulations have been introduced to model
uncertainty in the multiple-choice knapsack problem.~\citet{10549845} studied this problem and proposed a data-driven adaptive local search framework based on
sample-driven feasibility evaluation. However, their approach is restricted to
static and single-objective settings. 

Several studies investigated chance-constrained and dynamic variants of combinatorial optimization problems using MOEAs, including knapsack problem with stochastic weights and profits, monotone submodular functions, travelling thief problem and minimum spanning tree problem in the literature~\citep{10.1145/3377930.3390162,10.1145/3638529.3654066,DBLP:conf/gecco/0001W23, 10.1162/evco_a_00360,DBLP:conf/gecco/YanN024,DBLP:conf/cec/DonN025, DBLP:conf/gecco/DonN024,10.1007/978-3-031-70055-2_8}. Moreover, effectiveness of MOEAs for stochastic and dynamic optimization problems investigated as follows.~\citet{Roostapour2020} examined both single and multi-objective baseline EAs in the context of the dynamic classical knapsack problem, where the knapsack capacity changes over time. They showed that the multi-objective approach generally outperforms the single-objective algorithm, particularly when the frequency of dynamic changes is not excessively high. However, there are only a few studies that focus on using MOEAs to solve dynamic chance-constrained optimization problems.~\citet{DBLP:conf/ecai/AssimiHXN020} proposed a bi-objective formulation to address both the uncertain and dynamic aspects of the knapsack problem, where the weight of each item is independent and uniformly distributed. They have shown that the bi-objective optimization is particularly effective in managing dynamic chance-constrained problems. This work was further extended in~\citep{10.1145/3638529.3654067} by introducing 3-objective problem formulation, which was able to model the problem independently of the constraint’s required confidence level. They highlighted the effectiveness of 3-objective formulation over 2-objective formulation in addressing the dynamic chance-constrained knapsack problem. Moreover, ~\citet{10.1145/3638529.3654081} studied the 2-objective and 3-objective formulations of the dynamic variant of chance-constrained knapsack problem where the profit of each item is stochastic using GSEMO with standard and sliding window parent selection. Their results demonstrate that the 3-objective formulation yields improved performance. These studies demonstrate the effectiveness of MOEAs for solving combinatorial optimization problems under stochastic and dynamic settings.

\subsection{Contribution}

To the best of our knowledge, this is the first study of the dynamic chance-constrained multiple knapsack problem using bi-objective dominance- and decomposition-based MOEAs under an exact
probabilistic reformulation of the capacity constraints.

In this paper, we consider a chance-constrained multiple knapsack problem in which item weights are modeled as independent normally distributed random variables, while profits remain deterministic. This assumption enables an exact reformulation of the capacity constraints, avoiding the use of tail bounds or sampling-based approximations. Despite this exact reformulation, the resulting problem remains NP-hard and involves non-linear constraints. We further study a dynamic setting in which the capacity bounds of a subset of knapsacks change over time.

We formulate this problem as a bi-objective optimization problem that captures the trade-off between maximizing total profit and minimizing the aggregate chance-constrained weight. Solutions are represented using an integer-based encoding in which items are assigned directly to knapsacks. This representation provides an alternative to standard binary encodings for the multiple knapsack problem and implicitly enforces reserve constraints by construction. To address the dynamic nature of the problem, we employ a simple adaptive mutation-rate mechanism that supports search adaptation when capacity changes occur and no feasible solutions are available.

We conduct a comparative experimental evaluation using four well-established MOEAs, namely MOEA/D, NSGA-II, NSGA-III, and SPEA2 including both decomposition- and dominance-based search paradigms. We analyze their behavior under varying levels of uncertainty and dynamic capacity changes, providing empirical insights into the performance of bi-objective MOEAs for chance-constrained multiple knapsack problems in dynamic environments.

The remainder of the paper is organized as follows. Section~\ref{Sec: preliminaries}
defines the classical multiple knapsack problem and briefly describes the MOEAs
considered in this study. Section~\ref{Sec: CC_MKP} presents the chance-constrained
formulation of the stochastic MKP and the
integer-based solution encoding representation. Section~\ref{Sec: Dynamic_MKP} introduces the
dynamic problem setting and adaptive mutation rate strategy.
Section~\ref{Sec: Experimental_Investigations} describes the experimental setting and results. Finally, Section~\ref{Sec: Conclusion} concludes the paper.

\section{Preliminaries}
\label{Sec: preliminaries}
In this section, we define the classical MKP and describe the MOEAs considered.

\subsection{Classical Multiple Knapsack Problem}
\label{Sec: Classical_MKP}

The classical multiple knapsack problem assigns $n$ items to $m$ knapsacks, each with a capacity $B_i$, $i \in \{1,\dots,m\}$. Each item $j$ has a profit $p_j$ and a weight $w_j$. The objective is to select $m$ disjoint subsets of items that maximize the total profit while satisfying the capacity constraints of all knapsacks~\citep{CACCHIANI2022105693}.

A solution is represented by a binary matrix
$X = \{x_{ij}\}$, where $x_{ij} \in \{0,1\}$ indicates whether item $j$ is assigned to knapsack $i$.
The total profit of a solution is
$p(X) = \sum_{i=1}^m \sum_{j=1}^n p_j x_{ij}$,
and the total weight of knapsack $i$ is
$w_i(X) = \sum_{j=1}^n w_j x_{ij}$.

The MKP can be formulated as follows:
\begin{eqnarray}
\textbf{Maximize} & f(X) = p(X),\\
\textbf{Subject to} & 
w_i(X) \leq B_i, \forall i, \label{cons_1_s}\\
&\sum_{i=1}^m x_{ij} \leq 1, \quad \forall j, \label{cons_2_s}\\
&x_{ij} \in \{0,1\}, \label{cons_3_s}
\end{eqnarray}
where the constraint~\eqref{cons_1_s} ensures that the total weight assigned to each knapsack does not exceed its capacity. Constraint~\eqref{cons_2_s} enforces that each item is assigned to at most one knapsack. Constraint~\eqref{cons_3_s} specifies the binary nature of the decision variables.

In the following sections, we extend this problem to a stochastic and dynamic variants and reformulate it as a bi-objective optimization problem.

\subsection{Multi-Objective Evolutionary Algorithms}
\label{Sec: Algorithms}
We use four MOEAs to investigate the performance of our stochastic and dynamic MKP. These algorithms belong to two categories: decomposition- and dominance-based MOEAs. MOEA/D uses a decomposition-based approach that optimizes multiple scalar subproblems simultaneously using neighborhood information, while NSGA-II, NSGA-III, and SPEA2 are dominance-based algorithms.

In MOEA/D~\cite{4358754}, each subproblem is defined by a weight vector and an aggregation function, and neighborhood relationships among subproblems are determined based on the distance between weight vectors. MOEA/D supports several aggregation methods, including weighted sum, Tchebycheff, and penalty boundary intersection, and we use the Tchebycheff method in this study. The scalar optimization problem for each subproblem is defined as
$\text{Minimize} \quad
g^{te}(x \mid \lambda, z^{*}) =
\max_{1 \leq i \leq m}
\left\{ \lambda_i \lvert f_i(x) - z_i^{*} \rvert \right\}$,
where the weight vector $\lambda = (\lambda_1,\ldots,\lambda_m)$ satisfies $\lambda_i \geq 0$ and $\sum_{i=1}^m \lambda_i = 1$, $z^{*} = (z_1^{*}, \ldots, z_m^{*})$ denotes the ideal point, and $f_i(x)$ denotes the $i$-th objective function value of solution $x$.

NSGA-II~\cite{Deb2002AFA} applies fast non-dominated sorting and crowding distance to maintain diversity of the population, where each solution is assigned a non-domination rank and a crowding distance value. NSGA-III~\citep{6600851} follows the same approach as NSGA-II. It combines the parent population $S$ and offspring population $Q$ into a temporary population $R = S \cup Q$. When only a subset of solutions from the last front can be selected, a reference-point-based niching mechanism is applied. Each solution is associated with the nearest reference point using perpendicular distance.
SPEA2~\citep{ziztler2002spea2} employs an external archive to guide the evolutionary search, where each solution is assigned a fitness value based on dominance strength and density estimation computed using the distance to the $k$-th nearest neighbor in the objective space.

\section{Chance Constrained Multiple Knapsack Problem}
\label{Sec: CC_MKP}
In this section, we introduce chance-constrained multiple knapsack problem,
denoted as CC-MKP. 
This problem is important in applications where item weights are uncertain and deterministic capacity constraints may lead to infeasible or suboptimal solutions. 
CC-MKP extends the classical MKP by maximizing the original profit objective while requiring
that knapsack capacity constraints are satisfied with high probability. 

We assume that the weight of each item
\( w_j \) is a random variable following an independent normal distribution
\( \mathcal{N}(\mu_j, \sigma_j^2) \), where \( \mu_j \) denotes the mean and
\( \sigma_j^2 \) the variance of the weight of item $j$. For a given knapsack \( i \), the total expected weight is defined as $\mu_i(X) = \sum_{j=1}^n \mu_j x_{ij}$, and the corresponding variance is given by
$v_i(X) = \sum_{j=1}^n \sigma_j^2 x_{ij}$.

The chance constrained MKP is formulated as:
\begin{eqnarray}
\textbf{Maximize} & f(X)=p(X), \\
\textbf{Subject to} & \Pr\big(w_i(X) \leq B_i\big) \geq \alpha, \quad \forall i, \label{cons_cc_2}\\
& \sum_{i=1}^m x_{ij} \leq 1, \quad \forall j,\label{cons_cc_3} \\
& x_{ij} \in \{0,1\},
\end{eqnarray}
where $\alpha$ denotes the
confidence level, i.e., the minimum required probability that each knapsack
capacity constraint is satisfied. Our objective is to maximize the total profit while satisfying the chance
constraints \eqref{cons_cc_2} and the assignment (reserve) constraint
\eqref{cons_cc_3}. We assume that $\alpha \in [\alpha_l,\alpha_h]$ with $\alpha_l>1/2$ and $\alpha_h<1$, since we are interested in solutions with high confidence.

\subsection{Bi-objective Fitness Function for CC-MKP}

We propose a Pareto optimization approach for the chance constrained MKP by considering probabilistic constraint satisfaction as an additional objective, resulting in a bi-objective optimization problem defined as
\begin{equation}
\tilde g_{2D}(X) = (\tilde f_1(X), \tilde f_2(X)),
\label{eq: 2d_cc}
\end{equation}
where \( \tilde f_1 \) is maximized and \( \tilde f_2 \) is minimized.

Since we assume item weights are independent and normally distributed, the
probabilistic constraint in Equation~\eqref{cons_cc_2} can be reformulated as shown in~\citep{ijcai2022p665}. Specifically,
\begin{equation}
\tilde w_i(X) = \mu_i(X) + K_{\alpha}\sqrt{v_i(X)},
\label{eq: chance_c_weight}
\end{equation}
where \( K_{\alpha} \) denotes the \( \alpha \)-quantile of the standard normal
distribution. The chance constraint is therefore equivalent to
\( \tilde w_i(X) \leq B_i \). This reformulation eliminates the need for approximation
methods such as tail bounds or sampling when evaluating feasibility. Note that
\( K_{\alpha} \) is undefined for \( \alpha = 1 \) due to the unbounded tail of
the normal distribution.

The two objectives are defined as:
\begin{align}
\tilde f_1(X) &=
\begin{cases}
p(X), & \text{if } \tilde w_i(X) \leq B_i,\ \forall i, \\
-\tilde e(X), & \text{otherwise},
\end{cases}
\label{eq: objective_1_3d_cc} \\
\tilde f_2(X) &=
\begin{cases}
\sum_{i=1}^m \tilde w_i(X), & \text{if } \tilde w_i(X) \leq B_i,\ \forall i, \\
M + \tilde e(X), & \text{otherwise}.
\end{cases}
\label{eq: objective_2_3d_cc}
\end{align}
where \( p(X) \) denotes the total profit and \( M \) is a large
constant. We apply a penalty term $\tilde e(X) = \sum_{i=1}^m \max\big(0, \tilde w_i(X) - B_i\big)$ for infeasible solutions, representing the aggregated violation of the chance-constrained capacity bounds, such that any infeasible solution is always worse than any feasible one. The negative sign in the infeasible case of \( \tilde f_1 \)
ensures consistent penalization with respect to profit maximization. 
This formulation balances profit maximization and probabilistic capacity satisfaction at a given confidence level, providing a framework for modeling uncertainty in CC-MKP.

\subsection{Integer-based Encoding Scheme for MKP}
\label{Sec: Encoding}

In the MKP formulation, a solution is represented by a binary matrix
\( X = \{x_{ij}\} \in \{0,1\}^{m \times n} \), where \( x_{ij} = 1 \) indicates that
item \( j \) is assigned to knapsack \( i \).
This representation requires \( m \times n \) binary decision variables and
explicitly enforces the constraint that each item can be assigned to at most one
knapsack.

We adopt an integer-based encoding that provides an equivalent but
more compact representation by using a single decision variable per item instead
of one variable per item–knapsack pair~\cite{biomimetics10050274}. 
A solution is represented by a vector
\( X = (x_1, \dots, x_n) \in \{0,1,\dots,m\}^n \), where each variable \( x_j \)
specifies the knapsack assignment of item \( j \).
Specifically, \( x_j = 0 \) indicates that item \( j \) is not assigned to any
knapsack, while \( x_j = i \) with \( 1 \leq i \leq m \) indicates that item
\( j \) is assigned to knapsack \( i \).

The integer encoding corresponds to the binary assignment variables as $x_{ij} = 1$ if $x_j = i$, and $0$ otherwise.
Compared to the binary matrix representation, the integer encoding reduces the
number of decision variables from \( m \times n \) to \( n \), resulting in a
lower-dimensional search space.
Moreover, constraint~\eqref{cons_cc_3} (each item assigned to at most one knapsack)
is satisfied by construction, since each variable \( x_j \) takes a single value
in \( \{0,1,\dots,m\} \).

While this encoding reduces the number of decision variables, it increases
the domain size of each variable, resulting in a lower-dimensional search
space with larger variable domains.

\section{Chance Constrained Multiple Knapsack Problem with Dynamic Capacity Bounds}
\label{Sec: Dynamic_MKP}

\begin{algorithm}[t!]
  \caption{Dynamic adaptive mutation rate strategy~(DAMRS).} 
  \label{alg: adaptive_mutation}
  \begin{algorithmic}[1]
    \STATE $S \gets \{X\}$ initial population 
    \STATE $P_m \gets P_{m, \text{initial}}$
    \STATE $T \gets$ threshold for feasible solutions
    \STATE \texttt{is\_adapt} $\leftarrow$ \texttt{False}
    \WHILE{termination criterion is not met}
        \IF{problem has changed}
            \STATE Re-evaluate the population $S$
            \STATE Reset mutation probability: $P_m \gets P_{m, \text{initial}}$.
            \STATE Set \texttt{is\_adapt} $\leftarrow$ \texttt{False}
            \IF{no feasible solution exists in $S$}
                \STATE Set \texttt{is\_adapt} $\leftarrow$ \texttt{True}
                \STATE Increase mutation probability: $P_m \gets 2 \cdot P_m$
            \ENDIF
        \ENDIF
      \IF{\texttt{is\_adapt} \textbf{is} \texttt{True}}
        \STATE $F \gets$ count of feasible solutions in $S$
        \IF{$F \geq T$}
          \STATE Reset mutation probability: $P_m \gets P_{m, \text{initial}}$
          \STATE Set \texttt{is\_adapt} $\leftarrow$ \texttt{False}
        \ENDIF
      \ENDIF
    \ENDWHILE
  \end{algorithmic}
\end{algorithm}

We introduce a dynamic variant of the chance-constrained MKP, denoted as DCC-MKP. In this variant, the capacity bounds of a
subset of knapsacks change during the optimization process. Specifically, we introduce environmental changes at fixed intervals of
$\tau$ iterations. 
At each change point, a subset of knapsacks is selected
uniformly at random, and the capacity of each selected knapsack $i$ is modified
by a multiplicative factor $r_i$. The factor $r_i$ is chosen uniformly at random from the interval
$[1-\eta,\,1+\eta]$, where $\eta$ controls the magnitude of change.
This dynamic setting is used to evaluate the behavior of bi-objective MOEAs when
chance-constrained capacity bounds change during the optimization process.

The capacity of knapsack $i$ at iteration $t$ is defined as
\begin{equation}
B_i^{(t)} = 
\begin{cases}
B_i^{(0)} & \text{if } t = 0, \\
r_i \cdot B_i^{(0)} & \text{if } t \bmod \tau = 0 \text{ and $i$ is selected}, \\
B_i^{(0)} & \text{if } t \bmod \tau = 0 \text{ and $i$ is not selected}, \\
B_i^{(t-1)} & \text{otherwise}.
\end{cases}
\end{equation}
where $B_i^{(0)}$ is the baseline capacity. At change points, capacities are reset relative to $B_i^{(0)}$ rather than updated from $B_i^{(t-1)}$, avoiding drift from repeated multiplicative updates.

After each change, the population $S$ is re-evaluated with respect to the updated
capacity bounds, and infeasible solutions are penalized according to their
constraint violations.
If no feasible solution exists in the population following a change, a simple
dynamic adaptive mutation rate strategy~(DAMRS)~\cite{10.1007/978-3-540-70928-2_60,10.1145/3524495,10.1145/3450218.3477305} is applied to increase
exploration~(Algorithm~\ref{alg: adaptive_mutation}). 
The algorithm temporarily increases the mutation probability $P_m$ and resets it
to its initial value once the number of feasible solutions $F$ exceeds a
predefined threshold $T$.

\section{Experimental Investigations}
\label{Sec: Experimental_Investigations}

In this section, we first describe the benchmark instances and experimental setup. We then evaluate the performance of decomposition- and dominance-based MOEAs on the proposed bi-objective formulations for the CC-MKP and the DCC-MKP.

\subsection{Benchmark Instances}

We use existing MKP benchmark instances\footnote{\url{https://site.unibo.it/operations-research/en/research/library-of-codes-and-instances-1}} that have been widely used for solving the classical MKP~\citep{DELLAMICO2019886,WANG2024121170,DETTI2021105210}. These benchmark instances were originally proposed by~\citet{fukunaga2011branch,KATAOKA2014440} and later organized and extended by~\citet{DELLAMICO2019886}.
The instances are categorized into four sets (FK1–FK4) with varying numbers of items and knapsacks, and include three profit--weight correlation classes: strongly correlated, weakly correlated, and uncorrelated. In our experiments, we select representative instances from FK1, FK3, and FK4, including strongly correlated and uncorrelated classes, covering a range of problem sizes and item-to-knapsack ratios, as summarized in Table~\ref{tab:benchmark}.

\begin{table}[!t]
    \centering
    \small
    \caption{Characteristics of 
    MKP benchmark instances. 
    FK4 includes six different $n/m$ ratios.}
    \label{tab:benchmark}
    \begin{adjustbox}{max width=1\textwidth}
    \begin{tabular}{lc|c|cccccc}
        \hline
        \textbf{Set} & FK1 & FK3 & \multicolumn{6}{c}{FK4} \\
        \hline
        Items ($n$)     & 100 & 300 & 300 & 225 & 240 & 375 & 300 & 500 \\
        Knapsacks ($m$) & 10  & 30  & 150 & 75  & 60  & 75  & 50  & 50  \\
        Ratio ($n/m$)   & 10  & 10  & 2   & 3   & 4   & 5   & 6   & 10  \\
        \hline
    \end{tabular}
    \end{adjustbox}
\end{table}

In our bi-objective formulation, we consider the original item weight $w_j$ as the mean weight $\mu_j$, which is uniformly distributed in the range $[10,1000]$. We follow the same profit–weight correlation classes in the original benchmark definitions. 
In the uncorrelated class, profits $p_j$ are sampled uniformly from $[10,1000]$, 
while in the strongly correlated class profits are defined as $p_j = \mu_j + 10$. Knapsack capacities are taken directly from the benchmark definition, with
$B_i$ for $i=1,\ldots,m-1$ selected uniformly from
$\left[0.4 \cdot \frac{\sum_{j=1}^n \mu_j}{m},\,0.6 \cdot \frac{\sum_{j=1}^n \mu_j}{m}\right]$,
and the final capacity $B_m$ adjusted so that the total capacity equals
$0.5 \cdot \sum_{j=1}^n \mu_j$.

\subsection{Experimental Setting}

To ensure a fair and consistent comparison, we closely follow the
experimental setup in~\citep{ijcai2022p665,10.1145/3638529.3654066}.
The item weight $w_j$ in the CC-MKP is assumed to be independent and
normally distributed as $\mathcal{N}(\mu_j,\sigma_j^2)$, where $\mu_j$
corresponds to the deterministic weight in the classical MKP. The
variance $\sigma_j^2$ is sampled uniformly from
$[a_k \mu_j, a_{k+1} \mu_j]$, with $(a_k,a_{k+1}) \in \{(0.5,2),(5,10)\}$,
representing increasing uncertainty levels denoted by $v_1$ and $v_2$.
Following prior studies~\cite{ijcai2022p665,10.1145/3638529.3654066},
we consider four chance-constraint confidence levels,
$\alpha \in \{1-10^{-k} \mid k \in \{2,4,6,8\}\}$.

In the dynamic setting, we introduce $\nu \in \{20, 50, 100\}$ dynamic changes.
At each change, we modify the capacity $B_i$ of a selected knapsack $i$
by a random factor $r_i \in [1-\eta,1+\eta]$, where
$\eta=0.2$ 
controls the magnitude of change.
We set the adaptive mutation threshold to $T = 0.3N$, where $N$ denotes the population size, and conduct all dynamic experiments under the variance setting $v_1$.

We compare four MOEAs: MOEA/D, NSGA-II, NSGA-III, and SPEA2 covering both decomposition- and dominance-based methods. All algorithms use integer polynomial mutation with a probability of $1/n$ and integer SBX crossover with a probability of $0.9$, with a population size of $100$ and a budget of $10^7$ fitness evaluations.
Dynamic experiments include a warm-up phase of $10^6$ evaluations before introducing
capacity changes. MOEA/D uses the Tchebycheff aggregation function with a neighborhood
size of $10\%$, neighborhood selection probability $1$, uniformly distributed weight
vectors, and replaces at most $1\%$ of solutions per generation. NSGA-II applies binary
tournament selection and the offspring population size is set equal to the population size. NSGA-III uses $100$ uniformly distributed reference directions for
the two-objective case, and SPEA2 follows a configuration similar to NSGA-II.

For the CC-MKP, we evaluate performance using the mean and standard deviation of
the best profit obtained from the Pareto front over $30$ independent runs. For the DCC-MKP, we compute the partial offline error $\epsilon$ before each
environmental change, which measures the deviation between the profit of the best
algorithmic solution $p(X)$ and the optimal profit $p(X^*)$ of the corresponding
static classical MKP. We obtain $p(X^*)$ using the Gurobi Optimizer~\citep{gurobi}
with a time limit of $10{,}000$ seconds and a relative MIP gap of $0.5\%$. If $X$
is feasible, we define $\epsilon = p(X^*) - p(X)$; otherwise, we set
$\epsilon = p(X^*) + E$, where $E$ denotes the minimum constraint violation.
We report overall performance as the average partial offline error
$\sum_{k=1}^{\nu} \epsilon_k / \nu$. We analyze the statistical significance using the Kruskal--Wallis test at the
$95\%$ confidence level, followed by the Bonferroni post-hoc test
\cite{Corder_Foreman_2014,Neumann2020}. Detailed statistical results are provided
in the Appendix A and B.

\subsection{Results for Chance Constrained MKP}

First, we evaluate the performance of four algorithms on the CC-MKP in terms of the best obtained profits.

\begin{figure}[!t]
    \centering
        \includegraphics[width=0.7\textwidth]{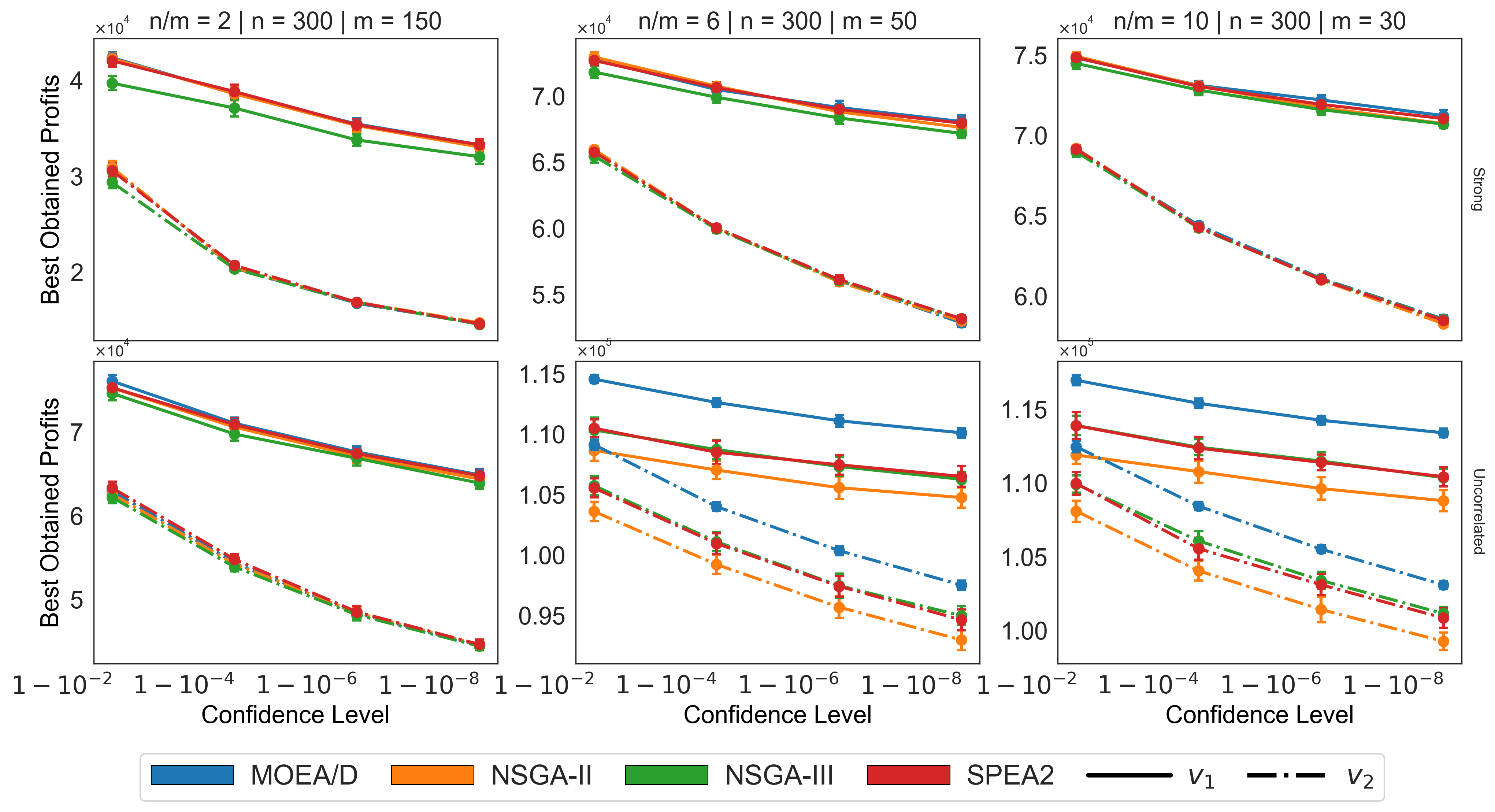}
    \caption{Mean and standard deviation of the best obtained profits for CC-MKP instances with a fixed number of items ($n=300$) and varying numbers of knapsacks ($m \in \{150,50,30\}$). 
    Results are shown for variance levels $v \in \{v_1,v_2\}$ across different confidence levels $\alpha$.}
    \label{fig: cc_mkp_n}
\end{figure}

Figure~\ref{fig: cc_mkp_n} illustrates algorithm performance across confidence levels for CC-MKP instances with a fixed number of items ($n=300$) and varying item-to-knapsack ratios ($n/m \in \{2,6,10\}$). Each row from top to bottom corresponds to strongly correlated and uncorrelated instances, respectively, while columns from left to right represent increasing ratios $n/m = 2, 6, 10$. The $y$-axis and $x$-axis show the best obtained profits over 30 runs and the confidence levels, respectively. Different colors indicate the MOEAs considered (MOEA/D~(blue), NSGA-II~(orange), NSGA-III~(green), and SPEA2~(red)), while line styles distinguish the two variance settings, with solid lines representing $v_1$ and dash-dotted lines representing $v_2$.

Performance consistently declines as variance and confidence
levels increase, since higher uncertainty and stricter chance constraints restrict item selection and force more risk-averse, lower-profit solutions. This effect increases with higher variance levels~($v_2$). 
Similar trends are observed across all the instances. For strongly correlated instances, all algorithms exhibit similar performance across all settings. In uncorrelated instances, when $n/m = 2$, the algorithms also perform comparably, as this configuration has greater flexibility in item selection. As the ratio $n/m$ increases, MOEA/D increasingly outperforms the other algorithms, while NSGA-II shows the worst performance. 

\begin{figure}[!t]
    \centering
    \includegraphics[width=0.7\textwidth]{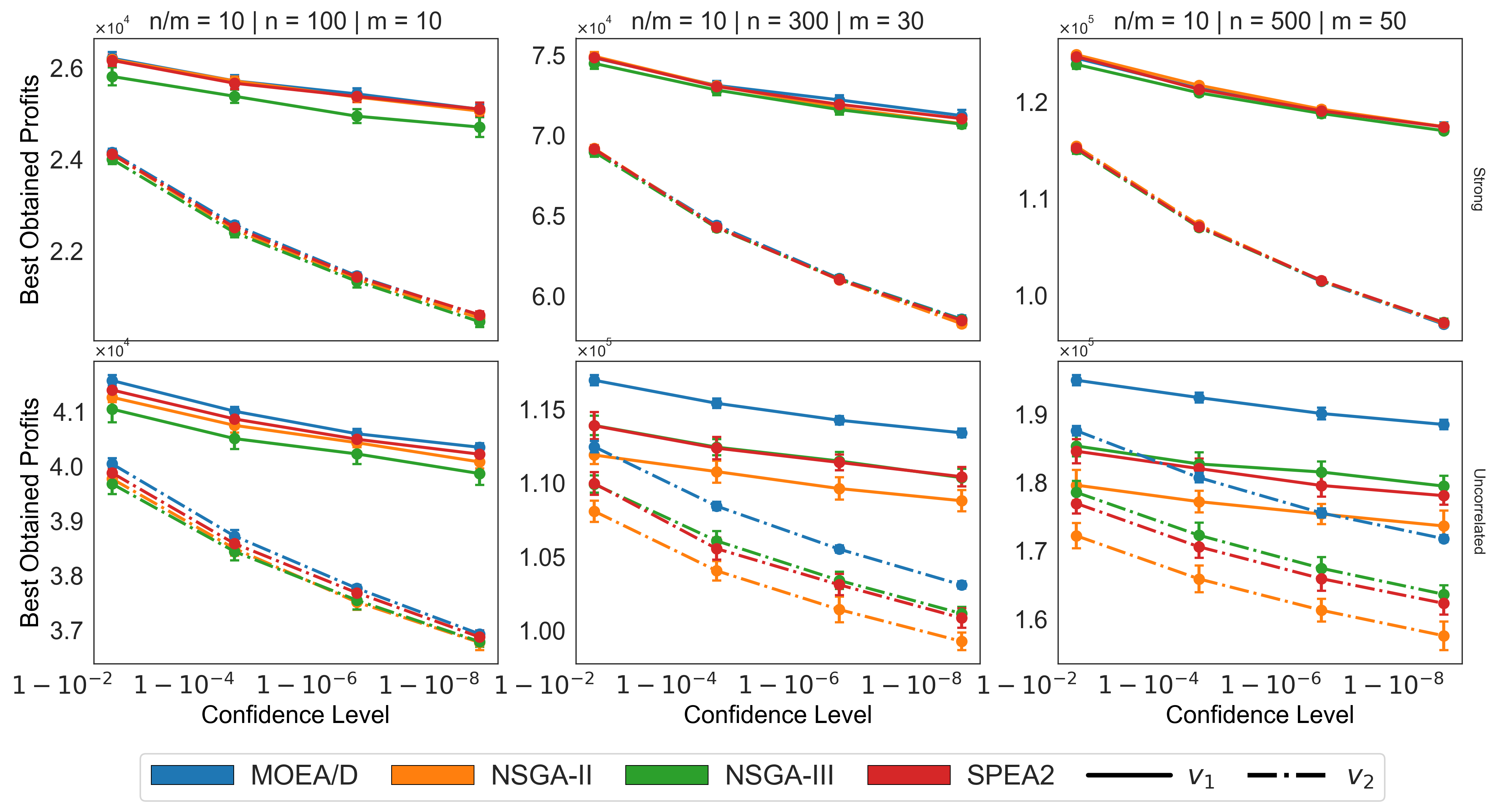}
    \caption{Mean and standard deviation of the best obtained profits for CC-MKP instances with a fixed item-to-knapsack ratio $n/m = 10$ and increasing problem sizes $(n,m) \in \{(100,10),(300,30),(500,50)\}$. Results are presented for variance levels $v \in \{v_1,v_2\}$ across different confidence levels $\alpha$.}
    \label{fig: cc_mkp_nm}
\end{figure}

Figure~\ref{fig: cc_mkp_nm} illustrates algorithm performance across confidence levels for CC-MKP instances with a fixed ratio $n/m = 10$. Columns correspond to $(n,m)\in\{(100,10),(300,30),(500,50)\}$, using the same layout as Figure~\ref{fig: cc_mkp_n}. For strongly correlated instances, all algorithms perform similarly at lower confidence levels, while MOEA/D outperforms as confidence level tighten. For uncorrelated instances, MOEA/D consistently outperforms the other algorithms across all variance and confidence levels, showing lower variability and higher profits. This advantage becomes more pronounced for larger instances~($n/m = 300/30$ and $500/50$), indicating better scalability under increasing problem size and uncertainty. 

\begin{figure*}[!t]
    \centering
    \includegraphics[width=\textwidth]{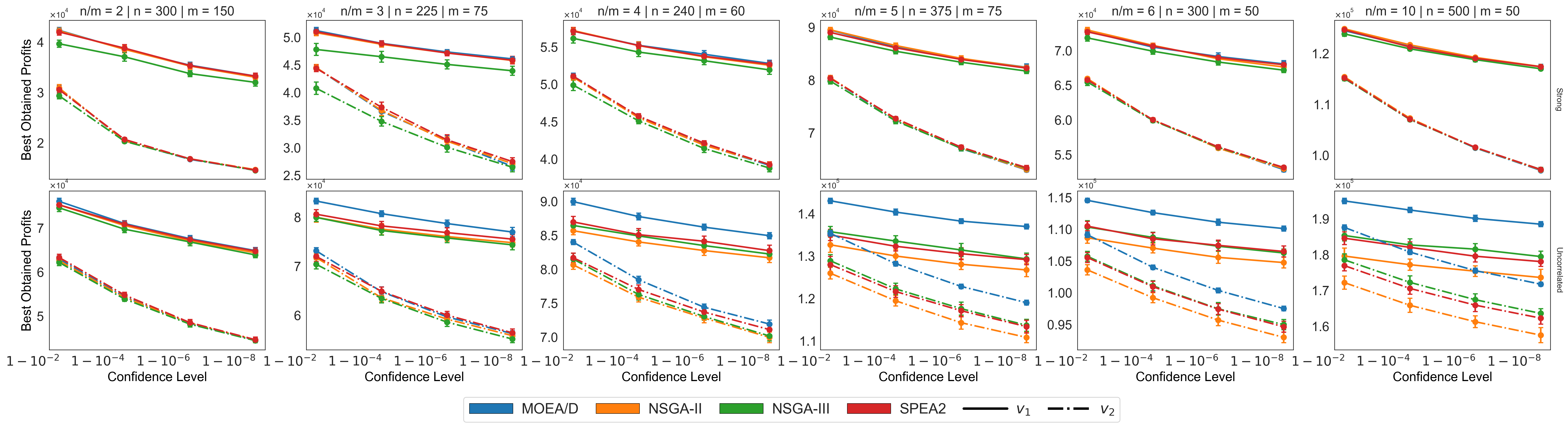}
    \caption{Mean and standard deviation of the best obtained profits for FK4 CC-MKP instances with varying item-to-knapsack ratios $n/m \in \{2,3,4,5,6,10\}$. Results are shown for variance levels $v \in \{v_1, v_2\}$ and four confidence levels $\alpha$.}
    \label{fig: cc_mkp_fk4}
\end{figure*}
Figure~\ref{fig: cc_mkp_fk4} presents results for FK4 instances with different item-to-knapsack ratios $n/m \in \{2,3,4,5,6,10\}$, shown from left to right in each subplot. All other parameters and settings are consistent with Figures~\ref{fig: cc_mkp_n} and~\ref{fig: cc_mkp_nm}. For strongly correlated instances, all algorithms exhibit similar performance. In contrast, for uncorrelated instances, MOEA/D consistently outperforms the other algorithms, particularly at higher $n/m$ ratios. NSGA-II shows the weakest performance, while NSGA-III and SPEA2 perform similarly but remain inferior to MOEA/D. Notably, for uncorrelated instances with $n/m = 10$, MOEA/D achieves higher profits at low confidence levels than those obtained by the other algorithms even under small variance setting~($v_1$). 

Overall, MOEA/D is effective for uncorrelated instances, particularly for larger $n/m$ ratios (e.g., $6$ or $10$), and performs comparably to other algorithms in strongly correlated instances. As $\alpha$ increases, solutions must move further away from the knapsack bound $B$ to satisfy stricter feasibility requirements, which reduces achievable profits. Similarly, increasing variance leads to lower profits because the algorithms must avoid high-risk items with uncertain weights. These effects are more pronounced for large, uncorrelated instances, where MOEA/D continues to outperform dominance-based algorithms, which degrade more rapidly. 

\subsection{Results for Dynamic Chance Constrained MKP}

Next, we analyze the performance of MOEAs on the DCC-MKP under dynamic capacity changes, considering different change frequencies and magnitudes in term of offline error. The lower offline error is, the better the performance is, as it indicates that the algorithm achieved a result closer to the optimal value. 

\begin{figure*}[!t]
    \centering
    \begin{subfigure}[b]{0.49\textwidth}
        \includegraphics[width=\textwidth]{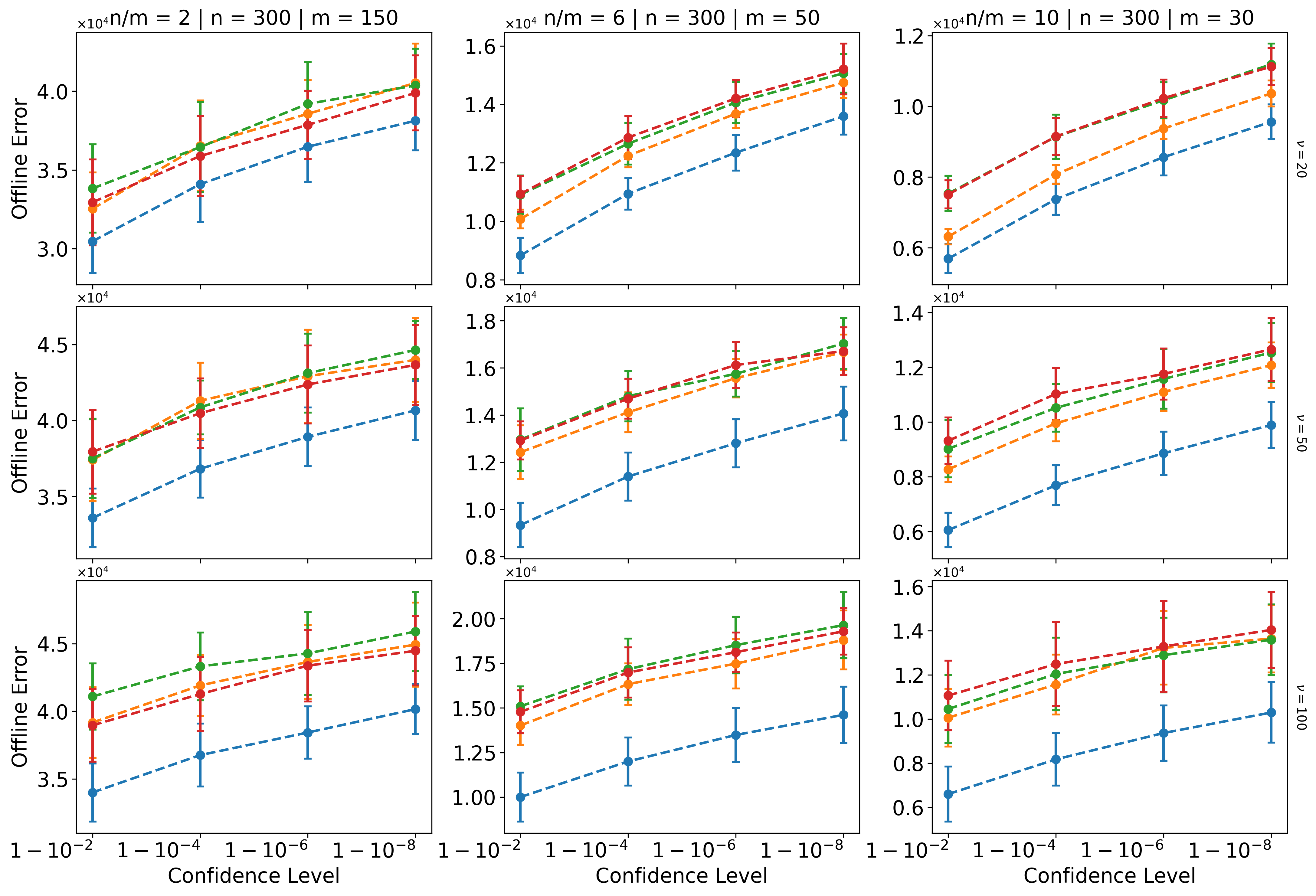}
        \caption{Type: strongly correlated, $\eta=0.2$}
    \end{subfigure}
    \centering
    \begin{subfigure}[b]{0.49\textwidth}
        \includegraphics[width=\textwidth]{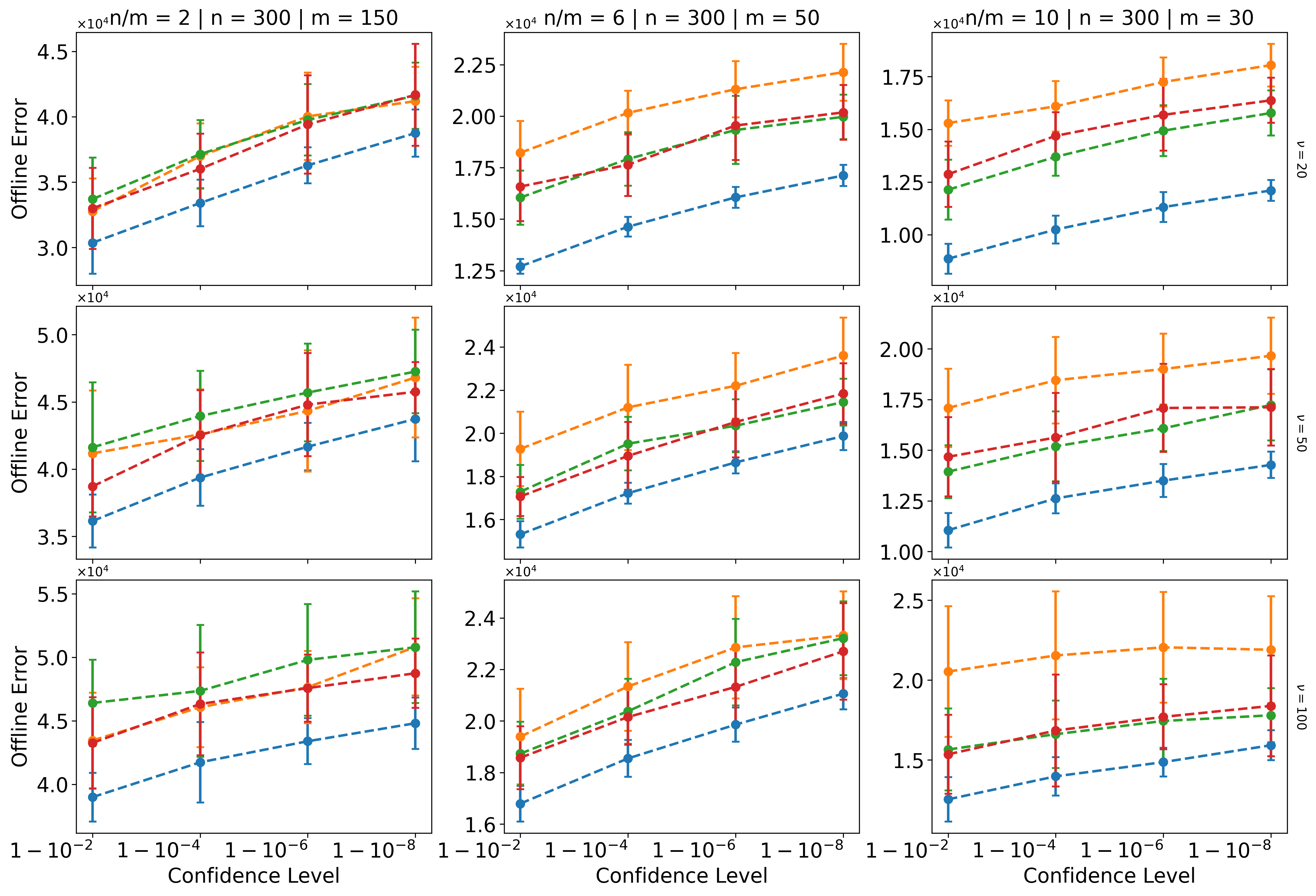}
        \caption{Type: uncorrelated, $\eta=0.2$}
    \end{subfigure}
    \centering
    \begin{subfigure}[b]{0.3\textwidth}
    \centering
    \includegraphics[width=\textwidth]{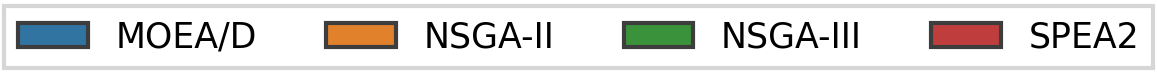}
    \end{subfigure}
    \caption{Mean and standard deviation of the offline error for DCC-MKP instances with a fixed number of items ($n=300$) and varying numbers of knapsacks~($m \in \{150, 50, 30\}$), under dynamic capacity changes with frequencies $\nu \in \{20, 50, 100\}$. Results are presented for strongly correlated and uncorrelated instances with change magnitude $\eta = 0.2$.}
    \label{fig: dcc_mkp_n}
\end{figure*}

\begin{figure*}[!t]
    \centering
    \begin{subfigure}[b]{0.49\textwidth}
        \includegraphics[width=\textwidth]{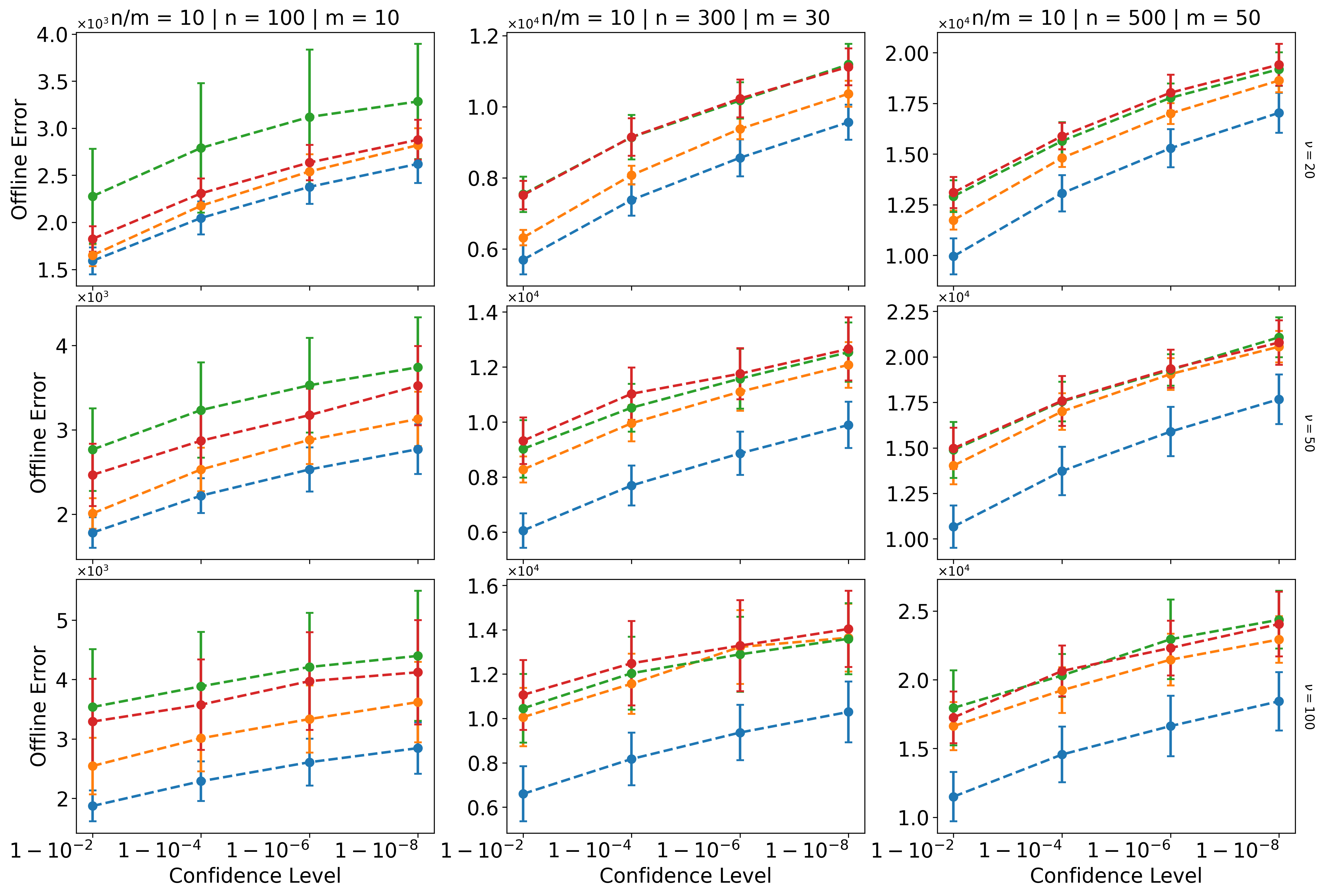}
        \caption{Type: strongly correlated, $\eta=0.2$}
    \end{subfigure}
    \centering
    \begin{subfigure}[b]{0.49\textwidth}
        \includegraphics[width=\textwidth]{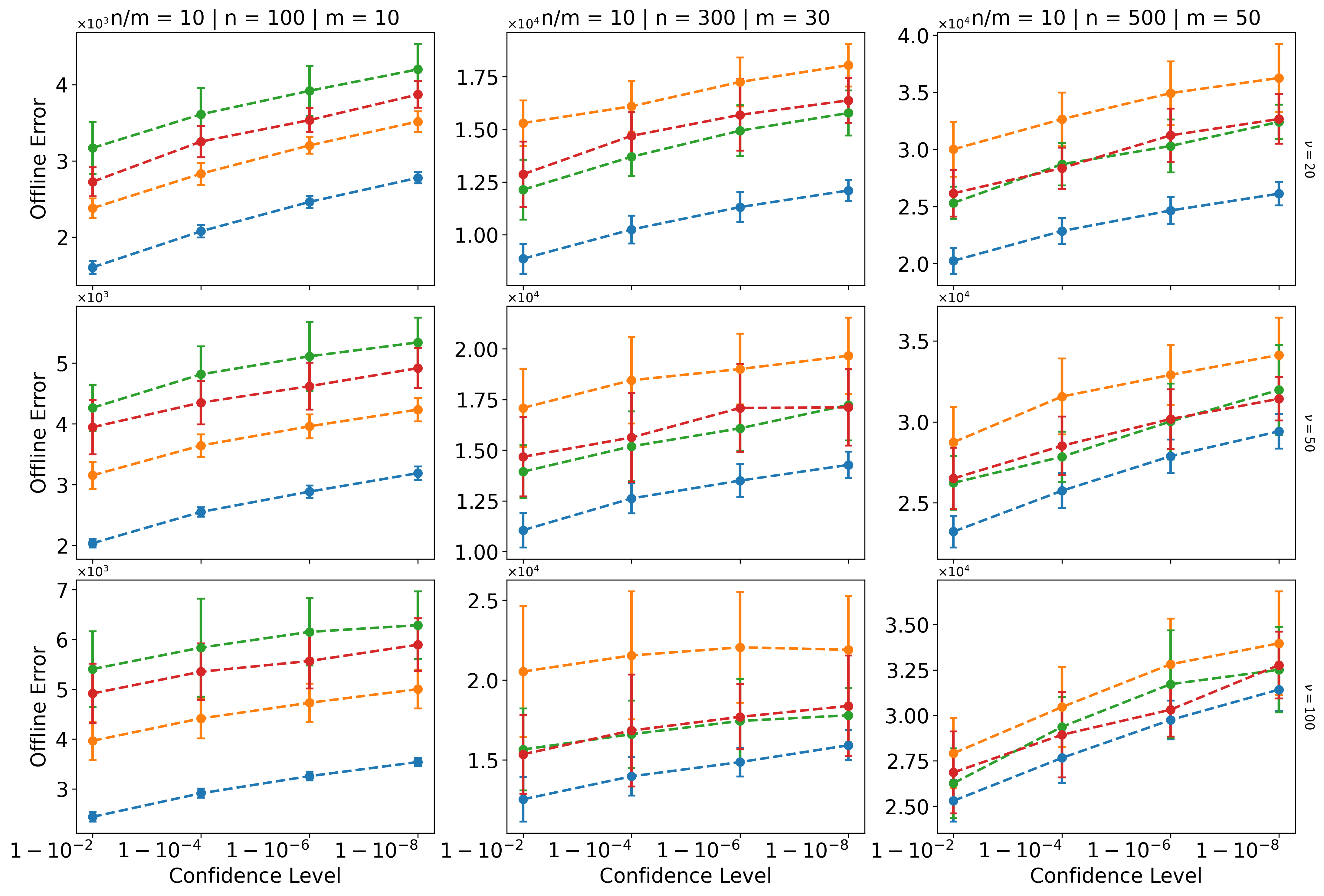}
        \caption{Type: uncorrelated, $\eta=0.2$}
    \end{subfigure}
    \centering
    \begin{subfigure}[b]{0.3\textwidth}
    \centering
    \includegraphics[width=\textwidth]{Figures/ledgend1.png}
    \end{subfigure}
    \caption{Mean and standard deviation of the offline error for DCC-MKP instances with a fixed item-to-knapsack ratio $n/m = 10$ and increasing problem sizes $(n,m) \in \{(100,10),(300,30),(500,50)\}$. Results are reported under dynamic capacity changes $\nu \in \{20,50,100\}$ for strongly correlated and uncorrelated instances with $\eta=0.2$.}
    \label{fig: dcc_mkp_nm}
\end{figure*}

\begin{figure*}[!t]
    \centering
    \begin{subfigure}[b]{0.96\textwidth}
        \includegraphics[width=\textwidth]{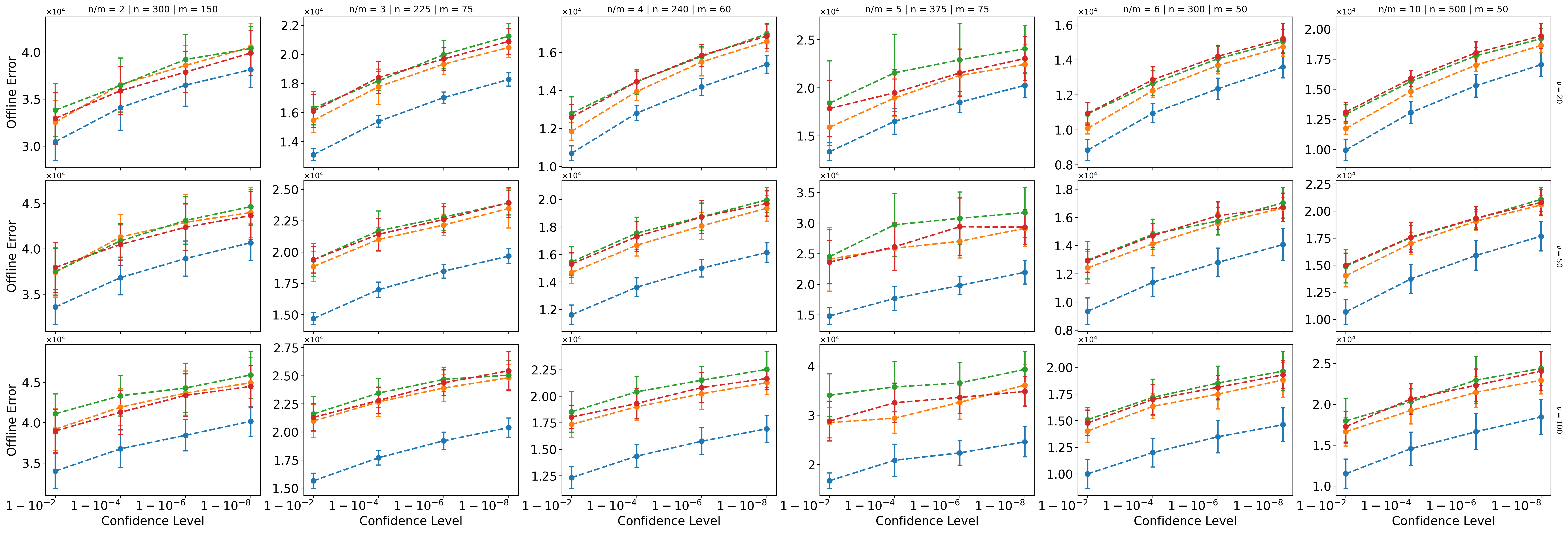}
        \caption{Type: strongly correlated}
    \end{subfigure}
    \centering
    \begin{subfigure}[b]{0.96\textwidth}
        \includegraphics[width=\textwidth]{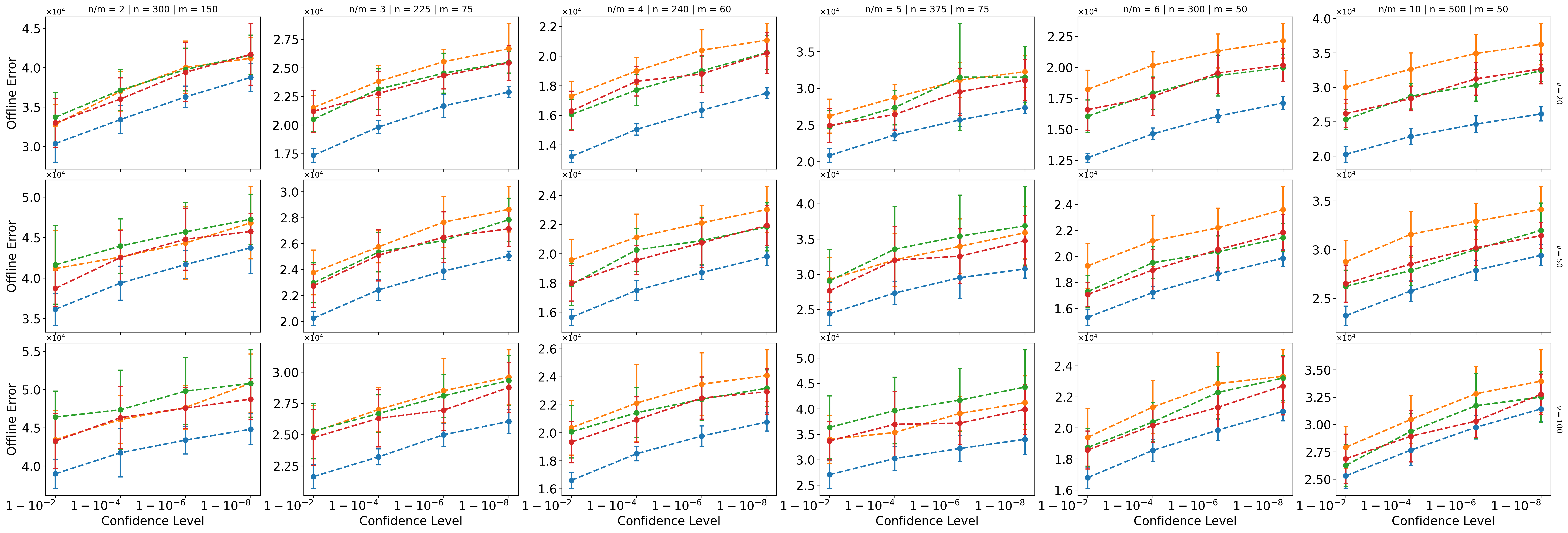}
        \caption{Type: uncorrelated}
    \end{subfigure}
    \begin{subfigure}[b]{0.4\textwidth}
    \centering
    \includegraphics[width=\textwidth]{Figures/ledgend1.png}
    \end{subfigure}
    \caption{Mean and standard deviation of the offline error for FK4 DCC-MKP instances with item-to-knapsack ratios $n/m \in \{2,3,4,5,6,10\}$. Results are shown under dynamic capacity changes $\nu \in \{20,50,100\}$ for strongly correlated and uncorrelated instances with $\eta=0.2$.}
    \label{fig: dcc_mkp_fk4_0.2}
\end{figure*}

Figure~\ref{fig: dcc_mkp_n} presents the mean and standard deviation of the offline error for strongly correlated and uncorrelated instances with $\eta = 0.2$. Each subplot presents results under three dynamic change settings, $\nu = 20, 50, 100$, with item-to-knapsack ratios ($n/m \in \{2,6,10\}$) and the number of items fixed at $n = 300$. Across all settings, the offline error increases as $\nu$ increases, reflecting the reduced time available for algorithms to adapt to changing capacity constraints. As $m$ decreases from $150$ to $30$, MOEA/D consistently outperforms the other algorithms. For strongly correlated instances, all dominance-based MOEAs perform comparably. For uncorrelated instances, NSGA-II exhibits the largest offline error, NSGA-III and SPEA2 perform similarly, and MOEA/D achieves the best results. 

Figure~\ref{fig: dcc_mkp_nm} presents results for instances with a fixed item-to-knapsack ratio $n/m = 10$ and increasing problem size ($n = 100- 500$). Each columns represent increasing problem sizes, following the same subplot organization as Figure~\ref{fig: dcc_mkp_n}. Although fixing $n/m$ maintains the same item density per knapsack, increasing $n$ expands the search space, leading to higher offline error under a fixed evaluation budget. This effect is further increases with $\nu$. Across most configurations, MOEA/D outperforms the other algorithms. For uncorrelated instances, NSGA-III exhibits the weakest performance for smaller instances, whereas NSGA-II yields the highest offline error for larger instances.

Figure~\ref{fig: dcc_mkp_fk4_0.2} summarizes performance for the FK4 instances across different item-to-knapsack ratios $n/m \in \{2, 3, 4, 5, 6, 10\}$ with $\eta = 0.2$.
The subplot organization follows the same layout as Figure~\ref{fig: dcc_mkp_n} and~\ref{fig: dcc_mkp_nm}.
As $n/m$ increases, the number of items per knapsack increases while the number of knapsacks
decreases, leading to higher within-knapsack combinatorial complexity. At low ratios ($n/m = 2, 3$), all algorithms exhibit comparable performance. As the ratio increases, MOEA/D’s advantage becomes more pronounced, achieving the
lowest offline error in almost all scenarios. 

Overall, MOEA/D consistently achieves best performance under dynamic conditions. Its decomposition-based framework and neighborhood update mechanism enable faster adaptation to environmental changes while preserving both convergence and diversity. In contrast, dominance-based algorithms exhibit slower recovery and large performance degradation as change frequency and problem complexity increase. 

\section{Conclusion}
\label{Sec: Conclusion}

In this paper, we investigated 
a stochastic and dynamic variant of the multiple knapsack problem with chance constraints, where item weights follow independent normal distribution and knapsack capacities change over time. We formulated the problem as a bi-objective optimization problem balancing profit maximization and probabilistic capacity satisfaction at a given confidence level. Experimental results show that increasing uncertainty and stricter confidence levels reduce achievable profits, while algorithm performance depends strongly on instance structure. MOEA/D consistently outperforms dominance-based approaches for uncorrelated instances, higher item-to-knapsack ratios, and under dynamic capacity changes, demonstrating faster adaptation and lower offline error. In contrast, dominance-based algorithms remain competitive mainly for strongly correlated and lower-dimensional instances. 
These findings highlight the effectiveness of bi-objective evolutionary optimization for the optimization problems under stochastic and dynamic conditions. Future work will extend the proposed framework to real-world application domains with uncertainty and dynamic constraints, such as mining production planning.

\section{Acknowledgement}
This work has been supported by the CNRS-Adelaide Mobility Scheme Award through grant 6021521.


\bibliographystyle{plainnat}
\bibliography{library} 

\appendix

\section{Analysis of the Performance of Chance Constrained MKP}
\label{app: cc_mkp}

Corresponding to Figures~1--3 in the main paper, Tables~\ref{tab: cc_results_1}--\ref{tab: cc_results_8} report the mean and standard deviation of the best obtained profits, along with statistical comparisons for the experimental results of the chance-constrained multiple knapsack problem. 

Each table corresponds to a specific benchmark instance. The columns are defined as follows: \textit{Set} denotes the instance set introduced in Table~1 of the main paper; $n/m$ represents the number of items ($n$) and knapsacks ($m$); \textit{Type} indicates the correlation between item profits and weights; $\alpha$ is the confidence level imposed on the chance constraint, with results reported for $\alpha \in \{1-10^{-2}, 1-10^{-4}, 1-10^{-6}, 1-10^{-8}\}$; and \textit{var} specifies the variance setting of each item ($v_1, v_2$). The subsequent columns correspond to the algorithms evaluated, MOEA/D, NSGA-II, NSGA-III, and SPEA2, where each algorithm includes subcolumns reporting the mean of the best obtained profits, the corresponding standard deviation, and the results of statistical comparisons. 

For statistical analysis, the Kruskal--Wallis test is applied at a $95\%$ confidence level, and when significant differences are detected, the Bonferroni post-hoc procedure is used for pairwise comparisons against a control algorithm. In the notation used, $X^{+}$ indicates that the algorithm in the given column significantly outperforms algorithm $X$, $X^{-}$ indicates that algorithm $X$ significantly outperforms the algorithm in the given column, and the absence of a symbol denotes no statistically significant difference between the compared algorithms.

\begin{table*}[!hbt]
\centering
\caption{Mean, standard deviation and statistical results of best obtained profits over 30 runs for CC-MKP with FK1\_100\_10 instance.}
\label{tab: cc_results_1}
\begin{adjustbox}{max width=1\textwidth}
\begin{tabular}{lllllrrrrrrrrrrrr}
\hline
 \multirow{2}{*}{\textbf{Set}} & \multirow{2}{*}{\textbf{n/m}} &  \multirow{2}{*}{\textbf{Type}} &  \multirow{2}{*}{\textbf{$\alpha$}} &  \multirow{2}{*}{\textbf{var}}  & \multicolumn{3}{c}{\textbf{MOEAD\textsuperscript{(1)}}} & \multicolumn{3}{c}{\textbf{NSGAII\textsuperscript{(2)}}} & \multicolumn{3}{c}{\textbf{NSGAIII\textsuperscript{(3)}}} & \multicolumn{3}{c}{\textbf{SPEA2\textsuperscript{(4)}}}\\
 & & & & & \textbf{Mean} & \textbf{Std} & \textbf{Stat} & \textbf{Mean} & \textbf{Std} & \textbf{Stat} & \textbf{Mean} & \textbf{Std} & \textbf{Stat} & \textbf{Mean} & \textbf{Std} & \textbf{Stat} \\
\hline
FK1  & 100/10 & Strong &0.99 & $v_1$& \textbf{26209.90} & 135.44 & 2\textsuperscript{(*)}3\textsuperscript{(+)}4\textsuperscript{(*)} & 26182.87 & 79.23 & 1\textsuperscript{(*)}3\textsuperscript{(+)}4\textsuperscript{(*)} & 25812.13 & 197.02 & 1\textsuperscript{(-)}2\textsuperscript{(-)}4\textsuperscript{(-)} & 26163.70 & 123.95 & 1\textsuperscript{(*)}2\textsuperscript{(*)}3\textsuperscript{(+)} \\
FK1  & 100/10 & Strong &0.99 & $v_2$& \textbf{24152.00} & 81.01 & 2\textsuperscript{(*)}3\textsuperscript{(+)}4\textsuperscript{(*)} & 24111.40 & 53.41 & 1\textsuperscript{(*)}3\textsuperscript{(+)}4\textsuperscript{(*)} & 24003.83 & 105.60 & 1\textsuperscript{(-)}2\textsuperscript{(-)}4\textsuperscript{(-)} & 24117.13 & 76.84 & 1\textsuperscript{(*)}2\textsuperscript{(*)}3\textsuperscript{(+)} \\
FK1  & 100/10 & Strong &0.9999 & $v_1$& \textbf{25715.77} & 124.53 & 2\textsuperscript{(*)}3\textsuperscript{(+)}4\textsuperscript{(*)} & 25714.07 & 93.63 & 1\textsuperscript{(*)}3\textsuperscript{(+)}4\textsuperscript{(*)} & 25379.87 & 145.77 & 1\textsuperscript{(-)}2\textsuperscript{(-)}4\textsuperscript{(-)} & 25664.40 & 129.50 & 1\textsuperscript{(*)}2\textsuperscript{(*)}3\textsuperscript{(+)} \\
\hline
\end{tabular}
\end{adjustbox}
\end{table*}

\begin{table*}[!hbt]
\centering
\caption{Mean, standard deviation and statistical results of best obtained profits over 30 runs for CC-MKP with FK3\_300\_30 instance.}
\label{tab: cc_results_2}
\begin{adjustbox}{max width=1\textwidth}\begin{tabular}{lllllrrrrrrrrrrrr}
\hline
 \multirow{2}{*}{\textbf{Set}} & \multirow{2}{*}{\textbf{n/m}} &  \multirow{2}{*}{\textbf{Type}} &  \multirow{2}{*}{\textbf{$\alpha$}} &  \multirow{2}{*}{\textbf{var}}  & \multicolumn{3}{c}{\textbf{MOEAD\textsuperscript{(1)}}} & \multicolumn{3}{c}{\textbf{NSGAII\textsuperscript{(2)}}} & \multicolumn{3}{c}{\textbf{NSGAIII\textsuperscript{(3)}}} & \multicolumn{3}{c}{\textbf{SPEA2\textsuperscript{(4)}}}\\
 & & & & & \textbf{Mean} & \textbf{Std} & \textbf{Stat} & \textbf{Mean} & \textbf{Std} & \textbf{Stat} & \textbf{Mean} & \textbf{Std} & \textbf{Stat} & \textbf{Mean} & \textbf{Std} & \textbf{Stat} \\
\hline
FK3 & 300/30 & Strong &0.99 & $v_1$& 74826.37 & 259.18 & 2\textsuperscript{(*)}3\textsuperscript{(+)}4\textsuperscript{(*)} & \textbf{74926.37} & 263.60 & 1\textsuperscript{(*)}3\textsuperscript{(+)}4\textsuperscript{(*)} & 74482.63 & 336.90 & 1\textsuperscript{(-)}2\textsuperscript{(-)}4\textsuperscript{(-)} & 74851.23 & 238.86 & 1\textsuperscript{(*)}2\textsuperscript{(*)}3\textsuperscript{(+)} \\
FK3 & 300/30 & Strong &0.99 & $v_2$& 69096.07 & 196.14 & 2\textsuperscript{(*)}3\textsuperscript{(*)}4\textsuperscript{(*)} & \textbf{69184.73} & 159.83 & 1\textsuperscript{(*)}3\textsuperscript{(+)}4\textsuperscript{(*)} & 68974.33 & 295.48 & 1\textsuperscript{(*)}2\textsuperscript{(-)}4\textsuperscript{(*)} & 69151.83 & 254.85 & 1\textsuperscript{(*)}2\textsuperscript{(*)}3\textsuperscript{(*)} \\
FK3 & 300/30 & Strong &0.9999 & $v_1$& \textbf{73114.07} & 274.32 & 2\textsuperscript{(*)}3\textsuperscript{(+)}4\textsuperscript{(*)} & 73085.03 & 187.44 & 1\textsuperscript{(*)}3\textsuperscript{(+)}4\textsuperscript{(*)} & 72822.97 & 313.63 & 1\textsuperscript{(-)}2\textsuperscript{(-)}4\textsuperscript{(*)} & 73043.77 & 252.45 & 1\textsuperscript{(*)}2\textsuperscript{(*)}3\textsuperscript{(*)} \\
FK3 & 300/30 & Strong &0.9999 & $v_2$& \textbf{64412.77} & 194.19 & - & 64300.60 & 203.08 & - & 64242.40 & 215.97 & - & 64284.83 & 213.58 & - \\
\hline
\end{tabular}
\end{adjustbox}
\end{table*}

\begin{table*}[!hbt]
\centering
\caption{Mean, standard deviation and statistical results of best obtained profits over 30 runs for CC-MKP with FK4\_300\_150 instance.}
\label{tab: cc_results_3}
\begin{adjustbox}{max width=1\textwidth}\begin{tabular}{lllllrrrrrrrrrrrr}
\hline
 \multirow{2}{*}{\textbf{Set}} & \multirow{2}{*}{\textbf{n/m}} &  \multirow{2}{*}{\textbf{Type}} &  \multirow{2}{*}{\textbf{$\alpha$}} &  \multirow{2}{*}{\textbf{var}}  & \multicolumn{3}{c}{\textbf{MOEAD\textsuperscript{(1)}}} & \multicolumn{3}{c}{\textbf{NSGAII\textsuperscript{(2)}}} & \multicolumn{3}{c}{\textbf{NSGAIII\textsuperscript{(3)}}} & \multicolumn{3}{c}{\textbf{SPEA2\textsuperscript{(4)}}}\\
 & & & & & \textbf{Mean} & \textbf{Std} & \textbf{Stat} & \textbf{Mean} & \textbf{Std} & \textbf{Stat} & \textbf{Mean} & \textbf{Std} & \textbf{Stat} & \textbf{Mean} & \textbf{Std} & \textbf{Stat} \\
\hline
FK4  & 300/150 & Strong &0.99 & $v_1$& \textbf{42372.97} & 577.32 & 2\textsuperscript{(*)}3\textsuperscript{(+)}4\textsuperscript{(*)} & 42272.00 & 601.13 & 1\textsuperscript{(*)}3\textsuperscript{(+)}4\textsuperscript{(*)} & 39706.17 & 712.29 & 1\textsuperscript{(-)}2\textsuperscript{(-)}4\textsuperscript{(-)} & 42045.73 & 637.96 & 1\textsuperscript{(*)}2\textsuperscript{(*)}3\textsuperscript{(+)} \\
FK4  & 300/150 & Strong &0.99 & $v_2$& 30836.97 & 567.94 & 2\textsuperscript{(*)}3\textsuperscript{(+)}4\textsuperscript{(*)} & \textbf{30930.53} & 671.84 & 1\textsuperscript{(*)}3\textsuperscript{(+)}4\textsuperscript{(*)} & 29381.13 & 629.29 & 1\textsuperscript{(-)}2\textsuperscript{(-)}4\textsuperscript{(-)} & 30593.07 & 698.86 & 1\textsuperscript{(*)}2\textsuperscript{(*)}3\textsuperscript{(+)} \\
FK4  & 300/150 & Strong &0.9999 & $v_1$& 38592.93 & 610.21 & 2\textsuperscript{(*)}3\textsuperscript{(+)}4\textsuperscript{(*)} & 38600.30 & 700.04 & 1\textsuperscript{(*)}3\textsuperscript{(+)}4\textsuperscript{(*)} & 37115.77 & 876.33 & 1\textsuperscript{(-)}2\textsuperscript{(-)}4\textsuperscript{(-)} & \textbf{38832.60} & 702.36 & 1\textsuperscript{(*)}2\textsuperscript{(*)}3\textsuperscript{(+)} \\
\hline
\end{tabular}
\end{adjustbox}
\end{table*}

\begin{table*}[!hbt]
\centering
\caption{Mean, standard deviation and statistical results of best obtained profits over 30 runs for CC-MKP with FK4\_225\_75 instance.}
\label{tab: cc_results_4}
\begin{adjustbox}{max width=1\textwidth}\begin{tabular}{lllllllrrrrrrrrrrrr}
\hline
 \multirow{2}{*}{\textbf{Set}} & \multirow{2}{*}{\textbf{n/m}} &  \multirow{2}{*}{\textbf{Type}} &  \multirow{2}{*}{\textbf{$\alpha$}} &  \multirow{2}{*}{\textbf{var}}  & \multicolumn{3}{c}{\textbf{MOEAD\textsuperscript{(1)}}} & \multicolumn{3}{c}{\textbf{NSGAII\textsuperscript{(2)}}} & \multicolumn{3}{c}{\textbf{NSGAIII\textsuperscript{(3)}}} & \multicolumn{3}{c}{\textbf{SPEA2\textsuperscript{(4)}}}\\
 & & & & & \textbf{Mean} & \textbf{Std} & \textbf{Stat} & \textbf{Mean} & \textbf{Std} & \textbf{Stat} & \textbf{Mean} & \textbf{Std} & \textbf{Stat} & \textbf{Mean} & \textbf{Std} & \textbf{Stat} \\
\hline
FK4 & 225/75 & Strong &0.99 & $v_1$& \textbf{51221.90} & 584.20 & 2\textsuperscript{(*)}3\textsuperscript{(+)}4\textsuperscript{(*)} & 50850.13 & 542.41 & 1\textsuperscript{(*)}3\textsuperscript{(+)}4\textsuperscript{(*)} & 47807.57 & 1080.55 & 1\textsuperscript{(-)}2\textsuperscript{(-)}4\textsuperscript{(-)} & 51020.83 & 578.87 & 1\textsuperscript{(*)}2\textsuperscript{(*)}3\textsuperscript{(+)} \\
FK4 & 225/75 & Strong &0.99 & $v_2$& 44496.90 & 452.87 & 2\textsuperscript{(*)}3\textsuperscript{(+)}4\textsuperscript{(*)} & \textbf{44534.03} & 548.40 & 1\textsuperscript{(*)}3\textsuperscript{(+)}4\textsuperscript{(*)} & 40776.50 & 1112.03 & 1\textsuperscript{(-)}2\textsuperscript{(-)}4\textsuperscript{(-)} & 44391.57 & 605.24 & 1\textsuperscript{(*)}2\textsuperscript{(*)}3\textsuperscript{(+)} \\
FK4 & 225/75 & Strong &0.9999 & $v_1$& \textbf{48916.60} & 427.86 & 2\textsuperscript{(*)}3\textsuperscript{(+)}4\textsuperscript{(*)} & 48745.20 & 500.29 & 1\textsuperscript{(*)}3\textsuperscript{(+)}4\textsuperscript{(*)} & 46492.00 & 987.06 & 1\textsuperscript{(-)}2\textsuperscript{(-)}4\textsuperscript{(-)} & 48872.80 & 502.49 & 1\textsuperscript{(*)}2\textsuperscript{(*)}3\textsuperscript{(+)} \\
\hline
\end{tabular}
\end{adjustbox}
\end{table*}

\begin{table*}[!hbt]
\centering
\caption{Mean, standard deviation and statistical results of best obtained profits over 30 runs for CC-MKP with FK4\_375\_75 instance.}
\label{tab: cc_results_5}
\begin{adjustbox}{max width=1\textwidth}\begin{tabular}{lllllrrrrrrrrrrrr}
\hline
 \multirow{2}{*}{\textbf{Set}} & \multirow{2}{*}{\textbf{n/m}} &  \multirow{2}{*}{\textbf{Type}} &  \multirow{2}{*}{\textbf{$\alpha$}} &  \multirow{2}{*}{\textbf{var}}  & \multicolumn{3}{c}{\textbf{MOEAD\textsuperscript{(1)}}} & \multicolumn{3}{c}{\textbf{NSGAII\textsuperscript{(2)}}} & \multicolumn{3}{c}{\textbf{NSGAIII\textsuperscript{(3)}}} & \multicolumn{3}{c}{\textbf{SPEA2\textsuperscript{(4)}}}\\
 & & & & & \textbf{Mean} & \textbf{Std} & \textbf{Stat} & \textbf{Mean} & \textbf{Std} & \textbf{Stat} & \textbf{Mean} & \textbf{Std} & \textbf{Stat} & \textbf{Mean} & \textbf{Std} & \textbf{Stat} \\
\hline
FK4  & 375/75 & Strong &0.99 & $v_1$& 89467.83 & 446.48 & 2\textsuperscript{(*)}3\textsuperscript{(+)}4\textsuperscript{(+)} & \textbf{89517.30} & 458.09 & 1\textsuperscript{(*)}3\textsuperscript{(+)}4\textsuperscript{(+)} & 88139.27 & 482.89 & 1\textsuperscript{(-)}2\textsuperscript{(-)}4\textsuperscript{(-)} & 89030.73 & 480.33 & 1\textsuperscript{(-)}2\textsuperscript{(-)}3\textsuperscript{(+)} \\
FK4  & 375/75 & Strong &0.99 & $v_2$& 80353.30 & 337.83 & 2\textsuperscript{(*)}3\textsuperscript{(+)}4\textsuperscript{(*)} & \textbf{80468.23} & 388.59 & 1\textsuperscript{(*)}3\textsuperscript{(+)}4\textsuperscript{(*)} & 79830.27 & 543.29 & 1\textsuperscript{(-)}2\textsuperscript{(-)}4\textsuperscript{(-)} & 80350.13 & 522.30 & 1\textsuperscript{(*)}2\textsuperscript{(*)}3\textsuperscript{(+)} \\
FK4  & 375/75 & Strong &0.9999 & $v_1$& 86223.47 & 520.97 & 2\textsuperscript{(*)}3\textsuperscript{(+)}4\textsuperscript{(*)} & \textbf{86539.00} & 457.88 & 1\textsuperscript{(*)}3\textsuperscript{(+)}4\textsuperscript{(+)} & 85464.93 & 511.82 & 1\textsuperscript{(-)}2\textsuperscript{(-)}4\textsuperscript{(-)} & 86110.53 & 522.25 & 1\textsuperscript{(*)}2\textsuperscript{(-)}3\textsuperscript{(+)} \\
\hline
\end{tabular}
\end{adjustbox}
\end{table*}

\begin{table*}[!hbt]
\centering
\caption{Mean, standard deviation and statistical results of best obtained profits over 30 runs for CC-MKP with FK4\_300\_50 instance.}
\label{tab: cc_results_6}
\begin{adjustbox}{max width=1\textwidth}\begin{tabular}{lllllrrrrrrrrrrrr}
\hline
 \multirow{2}{*}{\textbf{Set}} & \multirow{2}{*}{\textbf{n/m}} &  \multirow{2}{*}{\textbf{Type}} &  \multirow{2}{*}{\textbf{$\alpha$}} &  \multirow{2}{*}{\textbf{var}}  & \multicolumn{3}{c}{\textbf{MOEAD\textsuperscript{(1)}}} & \multicolumn{3}{c}{\textbf{NSGAII\textsuperscript{(2)}}} & \multicolumn{3}{c}{\textbf{NSGAIII\textsuperscript{(3)}}} & \multicolumn{3}{c}{\textbf{SPEA2\textsuperscript{(4)}}}\\
 & & & & & \textbf{Mean} & \textbf{Std} & \textbf{Stat} & \textbf{Mean} & \textbf{Std} & \textbf{Stat} & \textbf{Mean} & \textbf{Std} & \textbf{Stat} & \textbf{Mean} & \textbf{Std} & \textbf{Stat} \\
\hline
FK4  & 300/50 & Strong &0.99 & $v_1$& 72775.77 & 358.34 & 2\textsuperscript{(*)}3\textsuperscript{(+)}4\textsuperscript{(*)} & \textbf{72966.40} & 375.01 & 1\textsuperscript{(*)}3\textsuperscript{(+)}4\textsuperscript{(*)} & 71827.13 & 454.97 & 1\textsuperscript{(-)}2\textsuperscript{(-)}4\textsuperscript{(-)} & 72702.90 & 363.10 & 1\textsuperscript{(*)}2\textsuperscript{(*)}3\textsuperscript{(+)} \\
FK4  & 300/50 & Strong &0.99 & $v_2$& 65788.60 & 295.35 & 2\textsuperscript{(*)}3\textsuperscript{(+)}4\textsuperscript{(*)} & \textbf{65958.60} & 243.64 & 1\textsuperscript{(*)}3\textsuperscript{(+)}4\textsuperscript{(*)} & 65475.07 & 511.35 & 1\textsuperscript{(-)}2\textsuperscript{(-)}4\textsuperscript{(*)} & 65782.43 & 281.17 & 1\textsuperscript{(*)}2\textsuperscript{(*)}3\textsuperscript{(*)} \\
FK4  & 300/50 & Strong &0.9999 & $v_1$& 70500.00 & 383.22 & 2\textsuperscript{(*)}3\textsuperscript{(+)}4\textsuperscript{(*)} & \textbf{70754.93} & 274.58 & 1\textsuperscript{(*)}3\textsuperscript{(+)}4\textsuperscript{(*)} & 69916.53 & 433.15 & 1\textsuperscript{(-)}2\textsuperscript{(-)}4\textsuperscript{(-)} & 70645.20 & 423.87 & 1\textsuperscript{(*)}2\textsuperscript{(*)}3\textsuperscript{(+)} \\
FK4  & 300/50 & Strong &0.9999 & $v_2$& 60011.67 & 256.54 &-  & 60019.00 & 287.79 & - & 59937.07 & 300.93 & - & \textbf{60035.37} & 211.25 & - \\
\hline
\end{tabular}
\end{adjustbox}
\end{table*}

\begin{table*}[!hbt]
\centering
\caption{Mean, standard deviation and statistical results of best obtained profits over 30 runs for CC-MKP with FK4\_240\_60 instance.}
\label{tab: cc_results_7}
\begin{adjustbox}{max width=1\textwidth}\begin{tabular}{lllllrrrrrrrrrrrr}
\hline
 \multirow{2}{*}{\textbf{Set}} & \multirow{2}{*}{\textbf{n/m}} &  \multirow{2}{*}{\textbf{Type}} &  \multirow{2}{*}{\textbf{$\alpha$}} &  \multirow{2}{*}{\textbf{var}}  & \multicolumn{3}{c}{\textbf{MOEAD\textsuperscript{(1)}}} & \multicolumn{3}{c}{\textbf{NSGAII\textsuperscript{(2)}}} & \multicolumn{3}{c}{\textbf{NSGAIII\textsuperscript{(3)}}} & \multicolumn{3}{c}{\textbf{SPEA2\textsuperscript{(4)}}}\\
 & & & & & \textbf{Mean} & \textbf{Std} & \textbf{Stat} & \textbf{Mean} & \textbf{Std} & \textbf{Stat} & \textbf{Mean} & \textbf{Std} & \textbf{Stat} & \textbf{Mean} & \textbf{Std} & \textbf{Stat} \\
\hline
FK4  & 240/60 & Strong &0.99 & $v_1$& 57159.00 & 384.58 & 2\textsuperscript{(*)}3\textsuperscript{(+)}4\textsuperscript{(*)} & \textbf{57194.73} & 419.90 & 1\textsuperscript{(*)}3\textsuperscript{(+)}4\textsuperscript{(*)} & 56150.30 & 630.60 & 1\textsuperscript{(-)}2\textsuperscript{(-)}4\textsuperscript{(-)} & 57114.67 & 478.28 & 1\textsuperscript{(*)}2\textsuperscript{(*)}3\textsuperscript{(+)} \\
FK4  & 240/60 & Strong &0.99 & $v_2$& \textbf{51095.90} & 397.26 & 2\textsuperscript{(*)}3\textsuperscript{(+)}4\textsuperscript{(*)} & 50949.00 & 415.22 & 1\textsuperscript{(*)}3\textsuperscript{(+)}4\textsuperscript{(*)} & 49894.30 & 734.58 & 1\textsuperscript{(-)}2\textsuperscript{(-)}4\textsuperscript{(-)} & 51077.90 & 300.02 & 1\textsuperscript{(*)}2\textsuperscript{(*)}3\textsuperscript{(+)} \\
FK4  & 240/60 & Strong &0.9999 & $v_1$& \textbf{55218.13} & 478.40 & 2\textsuperscript{(*)}3\textsuperscript{(+)}4\textsuperscript{(*)} & 55177.67 & 501.71 & 1\textsuperscript{(*)}3\textsuperscript{(+)}4\textsuperscript{(*)} & 54313.80 & 610.94 & 1\textsuperscript{(-)}2\textsuperscript{(-)}4\textsuperscript{(-)} & 55191.80 & 388.22 & 1\textsuperscript{(*)}2\textsuperscript{(*)}3\textsuperscript{(+)} \\
\hline
\end{tabular}
\end{adjustbox}
\end{table*}

\begin{table*}[!hbt]
\centering
\caption{Mean, standard deviation and statistical results of best obtained profits over 30 runs for CC-MKP with FK4\_500\_50 instance.}
\label{tab: cc_results_8}
\begin{adjustbox}{max width=1\textwidth}\begin{tabular}{lllllrrrrrrrrrrrr}
\hline
\multirow{2}{*}{\textbf{Set}} & \multirow{2}{*}{\textbf{n/m}} &  \multirow{2}{*}{\textbf{Type}} &  \multirow{2}{*}{\textbf{$\alpha$}} &  \multirow{2}{*}{\textbf{var}}  & \multicolumn{3}{c}{\textbf{MOEAD\textsuperscript{(1)}}} & \multicolumn{3}{c}{\textbf{NSGAII\textsuperscript{(2)}}} & \multicolumn{3}{c}{\textbf{NSGAIII\textsuperscript{(3)}}} & \multicolumn{3}{c}{\textbf{SPEA2\textsuperscript{(4)}}}\\
 & & & & & \textbf{Mean} & \textbf{Std} & \textbf{Stat} & \textbf{Mean} & \textbf{Std} & \textbf{Stat} & \textbf{Mean} & \textbf{Std} & \textbf{Stat} & \textbf{Mean} & \textbf{Std} & \textbf{Stat} \\
\hline
FK4 & 500/50 & Strong &0.99 & $v_1$& 124517.30 & 264.30 & 2\textsuperscript{(-)}3\textsuperscript{(+)}4\textsuperscript{(*)} & \textbf{124872.50} & 278.68 & 1\textsuperscript{(+)}3\textsuperscript{(+)}4\textsuperscript{(*)} & 123872.73 & 478.31 & 1\textsuperscript{(-)}2\textsuperscript{(-)}4\textsuperscript{(-)} & 124679.43 & 314.86 & 1\textsuperscript{(*)}2\textsuperscript{(*)}3\textsuperscript{(+)} \\
FK4 & 500/50 & Strong &0.99 & $v_2$& 115269.20 & 285.51 & 2\textsuperscript{(*)}3\textsuperscript{(*)}4\textsuperscript{(*)} & \textbf{115418.73} & 195.44 & 1\textsuperscript{(*)}3\textsuperscript{(+)}4\textsuperscript{(*)} & 115094.47 & 439.36 & 1\textsuperscript{(*)}2\textsuperscript{(-)}4\textsuperscript{(*)} & 115222.93 & 260.01 & 1\textsuperscript{(*)}2\textsuperscript{(*)}3\textsuperscript{(*)} \\
FK4 & 500/50 & Strong &0.9999 & $v_1$& 121393.60 & 338.77 & 2\textsuperscript{(-)}3\textsuperscript{(+)}4\textsuperscript{(*)} & \textbf{121719.93} & 255.98 & 1\textsuperscript{(+)}3\textsuperscript{(+)}4\textsuperscript{(+)} & 120937.57 & 331.81 & 1\textsuperscript{(-)}2\textsuperscript{(-)}4\textsuperscript{(-)} & 121299.30 & 286.43 & 1\textsuperscript{(*)}2\textsuperscript{(-)}3\textsuperscript{(+)} \\
\hline
\end{tabular}
\end{adjustbox}
\end{table*}

\newpage
\clearpage

\section{Analysis of the Performance of Chance Constrained MKP in Dynamic Environments}
\label{app: dcc_mkp}

Corresponding to Figures~4--6 in the main paper, Tables~\ref{tab: dcc_results_1}--\ref{tab: dcc_results_8} show the mean and standard deviation of the offline error, together with statistical comparisons for the experimental results of the dynamic chance-constrained multiple knapsack problem (DCC-MKP). Each table corresponds to a specific benchmark instance. The columns are defined as follows: \textit{Set} denotes the instance set as introduced in Table~1 of the main paper; $n/m$ represents the number of items ($n$) and knapsacks ($m$); \textit{Type} indicates the correlation between item profits and weights; \textit{var} specifies the variance setting of each item ($v_1$); $\eta$ denotes the controlling factor of the magnitude of dynamic change; $\nu$ represents the frequency of dynamic change; and $\alpha$ is the confidence level imposed on the chance constraint, with results showed for $\alpha \in \{1-10^{-2}, 1-10^{-4}, 1-10^{-6}, 1-10^{-8}\}$. The subsequent columns correspond to the algorithms evaluated MOEA/D, NSGA-II, NSGA-III, and SPEA2, where each algorithm includes subcolumns reporting the mean of the offline error, the corresponding standard deviation, and the results of statistical comparisons. The statistical comparison procedure among the algorithms is the same as described in Appendix~\ref{app: cc_mkp}.

\begin{table*}[!hbt]
\centering
\caption{Mean, standard deviation and statistical results of offline error over 30 runs for DCC-MKP with FK1\_100\_10 instance.}
\label{tab: dcc_results_1}
\begin{adjustbox}{max width=1\textwidth}
\begin{tabular}{llllllllrrrrrrrrrrrrr}
\hline
\multirow{2}{*}{\textbf{Set}} &\multirow{2}{*}{\textbf{n/m}} &  \multirow{2}{*}{\textbf{Type}} &  \multirow{2}{*}{\textbf{var}}  &  \multirow{2}{*}{\textbf{$\eta$}}&  \multirow{2}{*}{\textbf{$\nu$}} &\multirow{2}{*}{\textbf{$\alpha$}}  & \multicolumn{3}{c}{\textbf{MOEAD\textsuperscript{(1)}}} & \multicolumn{3}{c}{\textbf{NSGAII\textsuperscript{(2)}}} & \multicolumn{3}{c}{\textbf{NSGAIII\textsuperscript{(3)}}} & \multicolumn{3}{c}{\textbf{SPEA2\textsuperscript{(4)}}} \\
& & & & & & & \textbf{Mean} & \textbf{Std} & \textbf{Stat} & \textbf{Mean} & \textbf{Std} & \textbf{Stat} & \textbf{Mean} & \textbf{Std} & \textbf{Stat} & \textbf{Mean} & \textbf{Std} & \textbf{Stat} &\\
\hline
FK1&100/10 & strong & $v_1$   & 0.2 & $100$ &0.99 &  \textbf{1870.91} & 259.82 & 2\textsuperscript{(+)}3\textsuperscript{(+)}4\textsuperscript{(+)} & 2543.65 & 474.67 & 1\textsuperscript{(-)}3\textsuperscript{(+)}4\textsuperscript{(+)} & 3535.66 & 974.21 & 1\textsuperscript{(-)}2\textsuperscript{(-)}4\textsuperscript{(*)} & 3290.91 & 719.35 & 1\textsuperscript{(-)}2\textsuperscript{(-)}3\textsuperscript{(*)} \\
FK1&100/10 & strong & $v_1$   & 0.2 & $100$ &0.9999 &  \textbf{2287.90} & 334.89 & 2\textsuperscript{(+)}3\textsuperscript{(+)}4\textsuperscript{(+)} & 3010.72 & 555.70 & 1\textsuperscript{(-)}3\textsuperscript{(+)}4\textsuperscript{(+)} & 3884.43 & 916.79 & 1\textsuperscript{(-)}2\textsuperscript{(-)}4\textsuperscript{(*)} & 3575.48 & 759.54 & 1\textsuperscript{(-)}2\textsuperscript{(-)}3\textsuperscript{(*)} \\
FK1&100/10 & strong & $v_1$   & 0.2 & $100$ &0.999999 &  \textbf{2607.55} & 393.69 & 2\textsuperscript{(+)}3\textsuperscript{(+)}4\textsuperscript{(+)} & 3333.49 & 565.37 & 1\textsuperscript{(-)}3\textsuperscript{(+)}4\textsuperscript{(+)} & 4209.54 & 912.34 & 1\textsuperscript{(-)}2\textsuperscript{(-)}4\textsuperscript{(*)} & 3972.11 & 820.25 & 1\textsuperscript{(-)}2\textsuperscript{(-)}3\textsuperscript{(*)} \\
\hline
\end{tabular}
\end{adjustbox}
\end{table*}

\begin{table*}[!hbt]
\centering
\caption{Mean, standard deviation and statistical results of offline error over 30 runs for DCC-MKP with FK3\_300\_30 instance.}
\label{tab: dcc_results_2}
\begin{adjustbox}{max width=1\textwidth}
\begin{tabular}{lllllllrrrrrrrrrrrrr}
\hline
\multirow{2}{*}{\textbf{Set}} &\multirow{2}{*}{\textbf{n/m}} &  \multirow{2}{*}{\textbf{Type}} &  \multirow{2}{*}{\textbf{var}}  &  \multirow{2}{*}{\textbf{$\eta$}}&  \multirow{2}{*}{\textbf{$\nu$}} &\multirow{2}{*}{\textbf{$\alpha$}}  & \multicolumn{3}{c}{\textbf{MOEAD\textsuperscript{(1)}}} & \multicolumn{3}{c}{\textbf{NSGAII\textsuperscript{(2)}}} & \multicolumn{3}{c}{\textbf{NSGAIII\textsuperscript{(3)}}} & \multicolumn{3}{c}{\textbf{SPEA2\textsuperscript{(4)}}} \\
& & & & & & & \textbf{Mean} & \textbf{Std} & \textbf{Stat} & \textbf{Mean} & \textbf{Std} & \textbf{Stat} & \textbf{Mean} & \textbf{Std} & \textbf{Stat} & \textbf{Mean} & \textbf{Std} & \textbf{Stat} &\\
\hline
FK3&300/30 & strong & $v_1$   & 0.2 & $100$ &0.99 &  \textbf{6596.97} & 1241.79 & 2\textsuperscript{(+)}3\textsuperscript{(+)}4\textsuperscript{(+)} & 10057.81 & 1314.35 & 1\textsuperscript{(-)}3\textsuperscript{(*)}4\textsuperscript{(*)} & 10453.18 & 1551.18 & 1\textsuperscript{(-)}2\textsuperscript{(*)}4\textsuperscript{(*)} & 11062.01 & 1574.51 & 1\textsuperscript{(-)}2\textsuperscript{(*)}3\textsuperscript{(*)} \\
FK3&300/30 & strong & $v_1$   & 0.2 & $100$ &0.9999 &  \textbf{8170.58} & 1190.76 & 2\textsuperscript{(+)}3\textsuperscript{(+)}4\textsuperscript{(+)} & 11564.49 & 1357.07 & 1\textsuperscript{(-)}3\textsuperscript{(*)}4\textsuperscript{(*)} & 12038.35 & 1641.47 & 1\textsuperscript{(-)}2\textsuperscript{(*)}4\textsuperscript{(*)} & 12488.63 & 1903.22 & 1\textsuperscript{(-)}2\textsuperscript{(*)}3\textsuperscript{(*)} \\
FK3&300/30 & strong & $v_1$   & 0.2 & $100$ &0.999999 &  \textbf{9363.52} & 1252.28 & 2\textsuperscript{(+)}3\textsuperscript{(+)}4\textsuperscript{(+)} & 13223.73 & 1663.82 & 1\textsuperscript{(-)}3\textsuperscript{(*)}4\textsuperscript{(*)} & 12892.21 & 1691.92 & 1\textsuperscript{(-)}2\textsuperscript{(*)}4\textsuperscript{(*)} & 13285.14 & 2052.29 & 1\textsuperscript{(-)}2\textsuperscript{(*)}3\textsuperscript{(*)} \\
\hline
\end{tabular}
\end{adjustbox}
\end{table*}

\begin{table*}[!hbt]
\centering
\caption{Mean, standard deviation and statistical results of offline error over 30 runs for DCC-MKP with FK4\_225\_75 instance.}
\label{tab: dcc_results_3}
\begin{adjustbox}{max width=1\textwidth}
\begin{tabular}{lllllllrrrrrrrrrrrrr}
\hline
\multirow{2}{*}{\textbf{Set}} &\multirow{2}{*}{\textbf{n/m}} & \multirow{2}{*}{\textbf{Type}} & \multirow{2}{*}{\textbf{var}} & \multirow{2}{*}{\textbf{$\eta$}} & \multirow{2}{*}{\textbf{$\nu$}} & \multirow{2}{*}{\textbf{$\alpha$}} & \multicolumn{3}{c}{\textbf{MOEAD\textsuperscript{(1)}}} & \multicolumn{3}{c}{\textbf{NSGAII\textsuperscript{(2)}}} & \multicolumn{3}{c}{\textbf{NSGAIII\textsuperscript{(3)}}} & \multicolumn{3}{c}{\textbf{SPEA2\textsuperscript{(4)}}} \\
& & & & & & & Mean & Std & Stat & Mean & Std & Stat & Mean & Std & Stat & Mean & Std & Stat \\
\hline
FK4&225/75 & strong & $v_1$   & 0.2 & $100$ &0.99 &  \textbf{15641.97} & 678.52 & 2\textsuperscript{(+)}3\textsuperscript{(+)}4\textsuperscript{(+)} & 20961.62 & 1461.33 & 1\textsuperscript{(-)}3\textsuperscript{(*)}4\textsuperscript{(*)} & 21585.53 & 1566.69 & 1\textsuperscript{(-)}2\textsuperscript{(*)}4\textsuperscript{(*)} & 21286.85 & 1205.82 & 1\textsuperscript{(-)}2\textsuperscript{(*)}3\textsuperscript{(*)} \\
FK4&225/75 & strong & $v_1$   & 0.2 & $100$ &0.9999 &  \textbf{17694.46} & 647.15 & 2\textsuperscript{(+)}3\textsuperscript{(+)}4\textsuperscript{(+)} & 22662.75 & 1280.78 & 1\textsuperscript{(-)}3\textsuperscript{(*)}4\textsuperscript{(*)} & 23446.26 & 1298.07 & 1\textsuperscript{(-)}2\textsuperscript{(*)}4\textsuperscript{(*)} & 22791.26 & 1185.53 & 1\textsuperscript{(-)}2\textsuperscript{(*)}3\textsuperscript{(*)} \\
FK4&225/75 & strong & $v_1$   & 0.2 & $100$ &0.999999 &  \textbf{19204.34} & 777.40 & 2\textsuperscript{(+)}3\textsuperscript{(+)}4\textsuperscript{(+)} & 23900.87 & 1188.17 & 1\textsuperscript{(-)}3\textsuperscript{(*)}4\textsuperscript{(*)} & 24665.65 & 1084.30 & 1\textsuperscript{(-)}2\textsuperscript{(*)}4\textsuperscript{(*)} & 24360.18 & 1154.85 & 1\textsuperscript{(-)}2\textsuperscript{(*)}3\textsuperscript{(*)} \\
\hline
\end{tabular}
\end{adjustbox}
\end{table*}

\begin{table*}[!hbt]
\centering
\caption{Mean, standard deviation and statistical results of offline error over 30 runs for DCC-MKP with FK4\_240\_60 instance.}
\label{tab: dcc_results_4}
\begin{adjustbox}{max width=1\textwidth}
\begin{tabular}{lllllllrrrrrrrrrrrrr}
\hline
\multirow{2}{*}{\textbf{Set}} &\multirow{2}{*}{\textbf{n/m}} &  \multirow{2}{*}{\textbf{Type}} &  \multirow{2}{*}{\textbf{var}}  &  \multirow{2}{*}{\textbf{$\eta$}}&  \multirow{2}{*}{\textbf{$\nu$}} &\multirow{2}{*}{\textbf{$\alpha$}}  & \multicolumn{3}{c}{\textbf{MOEAD\textsuperscript{(1)}}} & \multicolumn{3}{c}{\textbf{NSGAII\textsuperscript{(2)}}} & \multicolumn{3}{c}{\textbf{NSGAIII\textsuperscript{(3)}}} & \multicolumn{3}{c}{\textbf{SPEA2\textsuperscript{(4)}}} \\
& & & & & & &\textbf{Mean} & \textbf{Std} & \textbf{Stat} & \textbf{Mean} & \textbf{Std} & \textbf{Stat} & \textbf{Mean} & \textbf{Std} & \textbf{Stat} & \textbf{Mean} & \textbf{Std} & \textbf{Stat} &\\
\hline
FK4&240/60 & strong & $v_1$   & 0.2 & $100$ &0.99 &  \textbf{12318.03} & 1019.19 & 2\textsuperscript{(+)}3\textsuperscript{(+)}4\textsuperscript{(+)} & 17337.37 & 1208.19 & 1\textsuperscript{(-)}3\textsuperscript{(*)}4\textsuperscript{(*)} & 18527.33 & 1918.20 & 1\textsuperscript{(-)}2\textsuperscript{(*)}4\textsuperscript{(*)} & 18032.59 & 1115.01 & 1\textsuperscript{(-)}2\textsuperscript{(*)}3\textsuperscript{(*)} \\
FK4&240/60 & strong & $v_1$   & 0.2 & $100$ &0.9999 &  \textbf{14343.86} & 1085.82 & 2\textsuperscript{(+)}3\textsuperscript{(+)}4\textsuperscript{(+)} & 18999.63 & 1268.15 & 1\textsuperscript{(-)}3\textsuperscript{(+)}4\textsuperscript{(*)} & 20410.59 & 1423.61 & 1\textsuperscript{(-)}2\textsuperscript{(-)}4\textsuperscript{(*)} & 19306.72 & 1453.95 & 1\textsuperscript{(-)}2\textsuperscript{(*)}3\textsuperscript{(*)} \\
FK4&240/60 & strong & $v_1$   & 0.2 & $100$ &0.999999 &  \textbf{15758.36} & 1269.97 & 2\textsuperscript{(+)}3\textsuperscript{(+)}4\textsuperscript{(+)} & 20228.00 & 1481.00 & 1\textsuperscript{(-)}3\textsuperscript{(*)}4\textsuperscript{(*)} & 21508.36 & 1296.64 & 1\textsuperscript{(-)}2\textsuperscript{(*)}4\textsuperscript{(*)} & 20815.49 & 1445.98 & 1\textsuperscript{(-)}2\textsuperscript{(*)}3\textsuperscript{(*)} \\
\hline
\end{tabular}
\end{adjustbox}
\end{table*}

\begin{table*}[!hbt]
\centering
\caption{Mean, standard deviation and statistical results of offline error over 30 runs for DCC-MKP with FK4\_300\_150 instance.}
\label{tab: dcc_results_5}
\begin{adjustbox}{max width=1\textwidth}
\begin{tabular}{lllllllrrrrrrrrrrrrr}
\hline
\multirow{2}{*}{\textbf{Set}} &\multirow{2}{*}{\textbf{n/m}} &  \multirow{2}{*}{\textbf{Type}} &  \multirow{2}{*}{\textbf{var}}  &  \multirow{2}{*}{\textbf{$\eta$}}&  \multirow{2}{*}{\textbf{$\nu$}} &\multirow{2}{*}{\textbf{$\alpha$}}  & \multicolumn{3}{c}{\textbf{MOEAD\textsuperscript{(1)}}} & \multicolumn{3}{c}{\textbf{NSGAII\textsuperscript{(2)}}} & \multicolumn{3}{c}{\textbf{NSGAIII\textsuperscript{(3)}}} & \multicolumn{3}{c}{\textbf{SPEA2\textsuperscript{(4)}}} \\
& & & & & & &\textbf{Mean} & \textbf{Std} & \textbf{Stat} & \textbf{Mean} & \textbf{Std} & \textbf{Stat} & \textbf{Mean} & \textbf{Std} & \textbf{Stat} & \textbf{Mean} & \textbf{Std} & \textbf{Stat} &\\
\hline
FK4&300/150 & strong & $v_1$   & 0.2 & $100$ &0.99 &  \textbf{33995.89} & 2147.65 & 2\textsuperscript{(+)}3\textsuperscript{(+)}4\textsuperscript{(+)} & 39168.69 & 2606.08 & 1\textsuperscript{(-)}3\textsuperscript{(*)}4\textsuperscript{(*)} & 41095.89 & 2458.39 & 1\textsuperscript{(-)}2\textsuperscript{(*)}4\textsuperscript{(*)} & 38963.88 & 2684.96 & 1\textsuperscript{(-)}2\textsuperscript{(*)}3\textsuperscript{(*)} \\
FK4&300/150 & strong & $v_1$   & 0.2 & $100$ &0.9999 &  \textbf{36757.37} & 2331.40 & 2\textsuperscript{(+)}3\textsuperscript{(+)}4\textsuperscript{(+)} & 41910.07 & 2259.45 & 1\textsuperscript{(-)}3\textsuperscript{(*)}4\textsuperscript{(*)} & 43323.56 & 2510.30 & 1\textsuperscript{(-)}2\textsuperscript{(*)}4\textsuperscript{(*)} & 41276.80 & 2729.45 & 1\textsuperscript{(-)}2\textsuperscript{(*)}3\textsuperscript{(*)} \\
FK4&300/150 & strong & $v_1$   & 0.2 & $100$ &0.999999 &  \textbf{38430.21} & 1931.62 & 2\textsuperscript{(+)}3\textsuperscript{(+)}4\textsuperscript{(+)} & 43653.08 & 2723.46 & 1\textsuperscript{(-)}3\textsuperscript{(*)}4\textsuperscript{(*)} & 44284.03 & 3057.96 & 1\textsuperscript{(-)}2\textsuperscript{(*)}4\textsuperscript{(*)} & 43376.67 & 2643.79 & 1\textsuperscript{(-)}2\textsuperscript{(*)}3\textsuperscript{(*)} \\
\hline
\end{tabular}
\end{adjustbox}
\end{table*}

\begin{table*}[!hbt]
\centering
\caption{Mean, standard deviation and statistical results of offline error over 30 runs for DCC-MKP with FK4\_300\_50 instance.}
\label{tab: dcc_results_6}
\begin{adjustbox}{max width=1\textwidth}
\begin{tabular}{lllllllrrrrrrrrrrrrr}
\hline
\multirow{2}{*}{\textbf{Set}} &\multirow{2}{*}{\textbf{n/m}} &  \multirow{2}{*}{\textbf{Type}} &  \multirow{2}{*}{\textbf{var}}  &  \multirow{2}{*}{\textbf{$\eta$}}&  \multirow{2}{*}{\textbf{$\nu$}} &\multirow{2}{*}{\textbf{$\alpha$}}  & \multicolumn{3}{c}{\textbf{MOEAD\textsuperscript{(1)}}} & \multicolumn{3}{c}{\textbf{NSGAII\textsuperscript{(2)}}} & \multicolumn{3}{c}{\textbf{NSGAIII\textsuperscript{(3)}}} & \multicolumn{3}{c}{\textbf{SPEA2\textsuperscript{(4)}}} \\
& & & & & & &\textbf{Mean} & \textbf{Std} & \textbf{Stat} & \textbf{Mean} & \textbf{Std} & \textbf{Stat} & \textbf{Mean} & \textbf{Std} & \textbf{Stat} & \textbf{Mean} & \textbf{Std} & \textbf{Stat} &\\
\hline
FK4&300/50 & strong & $v_1$   & 0.2 & $100$ &0.99 &  \textbf{9998.69} & 1369.39 & 2\textsuperscript{(+)}3\textsuperscript{(+)}4\textsuperscript{(+)} & 14011.72 & 1086.75 & 1\textsuperscript{(-)}3\textsuperscript{(+)}4\textsuperscript{(*)} & 15087.25 & 1125.17 & 1\textsuperscript{(-)}2\textsuperscript{(-)}4\textsuperscript{(*)} & 14775.61 & 1203.87 & 1\textsuperscript{(-)}2\textsuperscript{(*)}3\textsuperscript{(*)} \\
FK4&300/50 & strong & $v_1$   & 0.2 & $100$ &0.9999 &  \textbf{11993.76} & 1348.12 & 2\textsuperscript{(+)}3\textsuperscript{(+)}4\textsuperscript{(+)} & 16333.24 & 1155.50 & 1\textsuperscript{(-)}3\textsuperscript{(*)}4\textsuperscript{(*)} & 17173.22 & 1719.45 & 1\textsuperscript{(-)}2\textsuperscript{(*)}4\textsuperscript{(*)} & 16983.22 & 1410.00 & 1\textsuperscript{(-)}2\textsuperscript{(*)}3\textsuperscript{(*)} \\
FK4&300/50 & strong & $v_1$   & 0.2 & $100$ &0.999999 &  \textbf{13480.68} & 1519.08 & 2\textsuperscript{(+)}3\textsuperscript{(+)}4\textsuperscript{(+)} & 17481.06 & 1389.69 & 1\textsuperscript{(-)}3\textsuperscript{(*)}4\textsuperscript{(*)} & 18519.65 & 1584.50 & 1\textsuperscript{(-)}2\textsuperscript{(*)}4\textsuperscript{(*)} & 18123.61 & 1105.63 & 1\textsuperscript{(-)}2\textsuperscript{(*)}3\textsuperscript{(*)} \\
\hline
\end{tabular}
\end{adjustbox}
\end{table*}

\begin{table*}[!hbt]
\centering
\caption{Mean, standard deviation and statistical results of offline error over 30 runs for DCC-MKP with FK4\_375\_75 instance.}
\label{tab: dcc_results_7}
\begin{adjustbox}{max width=1\textwidth}
\begin{tabular}{lllllllrrrrrrrrrrrrr}
\hline
\multirow{2}{*}{\textbf{Set}} &\multirow{2}{*}{\textbf{n/m}} &  \multirow{2}{*}{\textbf{Type}} &  \multirow{2}{*}{\textbf{var}}  &  \multirow{2}{*}{\textbf{$\eta$}}&  \multirow{2}{*}{\textbf{$\nu$}} &\multirow{2}{*}{\textbf{$\alpha$}}  & \multicolumn{3}{c}{\textbf{MOEAD\textsuperscript{(1)}}} & \multicolumn{3}{c}{\textbf{NSGAII\textsuperscript{(2)}}} & \multicolumn{3}{c}{\textbf{NSGAIII\textsuperscript{(3)}}} & \multicolumn{3}{c}{\textbf{SPEA2\textsuperscript{(4)}}} \\
& & & & & & &\textbf{Mean} & \textbf{Std} & \textbf{Stat} & \textbf{Mean} & \textbf{Std} & \textbf{Stat} & \textbf{Mean} & \textbf{Std} & \textbf{Stat} & \textbf{Mean} & \textbf{Std} & \textbf{Stat} &\\
\hline
FK4&375/75 & strong & $v_1$   & 0.2 & $100$ &0.99 &  \textbf{16720.06} & 1575.42 & 2\textsuperscript{(+)}3\textsuperscript{(+)}4\textsuperscript{(+)} & 28515.89 & 3129.33 & 1\textsuperscript{(-)}3\textsuperscript{(+)}4\textsuperscript{(*)} & 34044.07 & 4330.68 & 1\textsuperscript{(-)}2\textsuperscript{(-)}4\textsuperscript{(-)} & 28809.13 & 4031.96 & 1\textsuperscript{(-)}2\textsuperscript{(*)}3\textsuperscript{(+)} \\
FK4&375/75 & strong & $v_1$   & 0.2 & $100$ &0.9999 &  \textbf{20867.63} & 3233.02 & 2\textsuperscript{(+)}3\textsuperscript{(+)}4\textsuperscript{(+)} & 29399.37 & 3042.79 & 1\textsuperscript{(-)}3\textsuperscript{(+)}4\textsuperscript{(*)} & 35741.89 & 5075.49 & 1\textsuperscript{(-)}2\textsuperscript{(-)}4\textsuperscript{(*)} & 32529.35 & 3976.50 & 1\textsuperscript{(-)}2\textsuperscript{(*)}3\textsuperscript{(*)} \\
FK4&375/75 & strong & $v_1$   & 0.2 & $100$ &0.999999 &  \textbf{22363.08} & 2509.52 & 2\textsuperscript{(+)}3\textsuperscript{(+)}4\textsuperscript{(+)} & 32617.00 & 3409.31 & 1\textsuperscript{(-)}3\textsuperscript{(*)}4\textsuperscript{(*)} & 36536.27 & 4138.82 & 1\textsuperscript{(-)}2\textsuperscript{(*)}4\textsuperscript{(*)} & 33608.27 & 3308.35 & 1\textsuperscript{(-)}2\textsuperscript{(*)}3\textsuperscript{(*)} \\
\hline
\end{tabular}
\end{adjustbox}
\end{table*}

\begin{table*}[!hbt]
\centering
\caption{Mean, standard deviation and statistical results of offline error over 30 runs for DCC-MKP with FK4\_500\_50 instance.}
\label{tab: dcc_results_8}
\begin{adjustbox}{max width=1\textwidth}
\begin{tabular}{lllllllrrrrrrrrrrrrr}
\hline
\multirow{2}{*}{\textbf{Set}} &\multirow{2}{*}{\textbf{n/m}} &  \multirow{2}{*}{\textbf{Type}} &  \multirow{2}{*}{\textbf{var}}  &  \multirow{2}{*}{\textbf{$\eta$}}&  \multirow{2}{*}{\textbf{$\nu$}} &\multirow{2}{*}{\textbf{$\alpha$}}  & \multicolumn{3}{c}{\textbf{MOEAD\textsuperscript{(1)}}} & \multicolumn{3}{c}{\textbf{NSGAII\textsuperscript{(2)}}} & \multicolumn{3}{c}{\textbf{NSGAIII\textsuperscript{(3)}}} & \multicolumn{3}{c}{\textbf{SPEA2\textsuperscript{(4)}}} \\
& & & & & & &\textbf{Mean} & \textbf{Std} & \textbf{Stat} & \textbf{Mean} & \textbf{Std} & \textbf{Stat} & \textbf{Mean} & \textbf{Std} & \textbf{Stat} & \textbf{Mean} & \textbf{Std} & \textbf{Stat} &\\
\hline
FK4&500/50 & strong & $v_1$   & 0.2 & $100$ &0.99 &  \textbf{11495.55} & 1801.93 & 2\textsuperscript{(+)}3\textsuperscript{(+)}4\textsuperscript{(+)} & 16627.49 & 1754.35 & 1\textsuperscript{(-)}3\textsuperscript{(*)}4\textsuperscript{(*)} & 17952.25 & 2729.89 & 1\textsuperscript{(-)}2\textsuperscript{(*)}4\textsuperscript{(*)} & 17252.88 & 1882.07 & 1\textsuperscript{(-)}2\textsuperscript{(*)}3\textsuperscript{(*)} \\
FK4&500/50 & strong & $v_1$   & 0.2 & $100$ &0.9999 &  \textbf{14560.17} & 2026.62 & 2\textsuperscript{(+)}3\textsuperscript{(+)}4\textsuperscript{(+)} & 19234.93 & 1660.58 & 1\textsuperscript{(-)}3\textsuperscript{(*)}4\textsuperscript{(*)} & 20309.30 & 1560.24 & 1\textsuperscript{(-)}2\textsuperscript{(*)}4\textsuperscript{(*)} & 20638.85 & 1842.09 & 1\textsuperscript{(-)}2\textsuperscript{(*)}3\textsuperscript{(*)} \\
FK4&500/50 & strong & $v_1$   & 0.2 & $100$ &0.999999 &  \textbf{16632.85} & 2196.45 & 2\textsuperscript{(+)}3\textsuperscript{(+)}4\textsuperscript{(+)} & 21459.60 & 1884.49 & 1\textsuperscript{(-)}3\textsuperscript{(*)}4\textsuperscript{(*)} & 22946.68 & 2896.75 & 1\textsuperscript{(-)}2\textsuperscript{(*)}4\textsuperscript{(*)} & 22310.20 & 1993.80 & 1\textsuperscript{(-)}2\textsuperscript{(*)}3\textsuperscript{(*)} \\
\hline
\end{tabular}
\end{adjustbox}
\end{table*}

\end{document}